%% file: main.tex
\documentclass[10pt,twocolumn,letterpaper]{article}

\usepackage[pagenumbers]{cvpr}   %

\input{preamble}

\newif\ifsuppinline
\suppinlinetrue

\definecolor{cvprblue}{rgb}{0.21,0.49,0.74}
\usepackage{indentfirst}   %
\usepackage[pagebackref,breaklinks,colorlinks,allcolors=cvprblue]{hyperref}
\usepackage{latexml}   %

\def\confName{}
\def\confYear{}

\title{Spectral Vision Transformer for Efficient Tokenization with Limited Data}

\iflatexml
\author{Alexandra G. Roberts \\ Cornell University \and
Maneesh John \\ Cornell University \and
Jinwei Zhang \\ University of Massachusetts Amherst \and
Dominick Romano \\ Cornell University \and
Mert \c{S}i\c{s}man \\ Cornell University \and
Ki Sueng Choi \\ Icahn School of Medicine at Mount Sinai \and
Heejong Kim \\ Weill Cornell Medicine \and
Mert R. Sabuncu \\ Cornell University \and
Thanh D. Nguyen \\ Weill Cornell Medicine \and
Alexey V. Dimov \\ Weill Cornell Medicine \and
Pascal Spincemaille \\ Cornell University \and
Brian H. Kopell \\ Icahn School of Medicine at Mount Sinai \and
Yi Wang \\ Weill Cornell Medicine}
\else
\author{Alexandra G. Roberts$^{1}$ \quad Maneesh John$^{1}$ \quad Jinwei Zhang$^{2}$ \quad Dominick Romano$^{1}$ \\
Mert \c{S}i\c{s}man$^{1}$ \quad Ki Sueng Choi$^{3}$ \quad Heejong Kim$^{4}$ \quad Mert R. Sabuncu$^{1}$ \quad Thanh D. Nguyen$^{4}$ \\
Alexey V. Dimov$^{4}$ \quad Pascal Spincemaille$^{1}$ \quad Brian H. Kopell$^{3}$ \quad Yi Wang$^{4}$ \\[4pt]
{\normalsize $^{1}$Cornell University \quad $^{2}$University of Massachusetts Amherst} \\
{\normalsize $^{3}$Icahn School of Medicine at Mount Sinai \quad $^{4}$Weill Cornell Medicine}}
\fi

\begin{document}
\twocolumn[{%
\renewcommand\twocolumn[1][]{#1}%
\maketitle
\begin{center}
  \captionsetup{type=figure}
  \includegraphics[width=\textwidth]{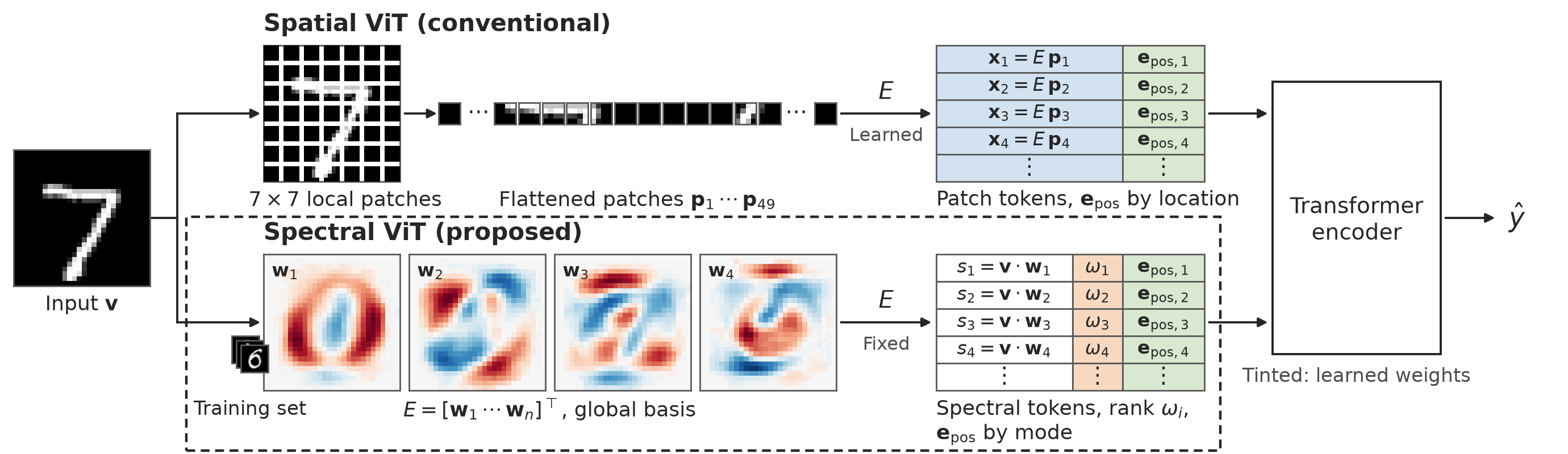}
  \captionof{figure}{Our proposed spectral ViT differs from conventional ViTs by constructing spectral rather than spatial tokens. Spectral tokens from any ordered basis possess inherent rank or hierarchical order. Our approach introduces a global inductive bias and spectral positional embeddings $\mathbf{e}_\mathrm{pos}$ by mode rather than local patch embedding. Tinted cells carry learned weights. The spatial ViT must learn its projection, whereas the spectral projection is fixed, leaving only the rank weights $\omega_i$ and the positional embeddings to be learned.} %
  \label{fig:schematic-diagram}
\end{center}%
\vspace{4pt}
}]
\input{sec/0_abstract} 
\input{sec/1_intro}
\input{sec/2_methods}
\input{sec/3_experiments}
\input{sec/4_results}
\input{sec/5_limitations}

\ifsuppinline\input{sec/X_suppl}\fi

{
    \small
    \bibliographystyle{ieeenat_fullname}
    \bibliography{references}
}

\end{document}

%% file: preamble.tex
\newcommand{\TODO}[1]{\textbf{\color{red}[TODO: #1]}}
\newcommand{\ed}[1]{{\color{red}\bfseries #1}}
\renewcommand{\TODO}[1]{}

\renewcommand{\ed}[1]{#1}

\usepackage[utf8]{inputenc}
\usepackage[T1]{fontenc}
\usepackage{amsfonts}
\usepackage{url}
\IfFileExists{placeins.sty}{\usepackage{placeins}}{}
\IfFileExists{tabularx.sty}{\usepackage{tabularx}}{}
\IfFileExists{boondox-cal.sty}{\usepackage{boondox-cal}}{}
\graphicspath{{fig/}}

\newenvironment{fequation}
{%
  \ifdefined\linenomath\begin{linenomath}\fi
  \begin{equation}
}
{%
  \end{equation}
  \ifdefined\linenomath\end{linenomath}\fi
}

\IfFileExists{stfloats.sty}{\usepackage{stfloats}\fnbelowfloat}{}

\DeclareRobustCommand{\secname}[1]{%
  \begingroup
  \renewcommand{\texorpdfstring}[2]{}%
  \nameref{#1}%
  \endgroup}

\input{suppnames}

\newcommand{\suppnumref}[1]{Appendix~\ref{#1}}

%% file: suppnames.tex
\expandafter\def\csname suppname@sec:archcmp\endcsname{Architecture comparison}
\expandafter\def\csname suppname@sec:ca\endcsname{Complexity analysis}
\expandafter\def\csname suppname@sec:encform\endcsname{Encoder form}
\expandafter\def\csname suppname@sec:basisprops\endcsname{Spectral basis properties}
\expandafter\def\csname suppname@sec:lra\endcsname{Rank-$n$ optimality}
\expandafter\def\csname suppname@sec:ti\endcsname{Translation invariance}
\expandafter\def\csname suppname@sec:tp\endcsname{Topology preservation}
\expandafter\def\csname suppname@sec:dctklt\endcsname{Eigenbasis approximation}
\expandafter\def\csname suppname@sec:corr\endcsname{Spatial correspondence}
\expandafter\def\csname suppname@sec:magcost\endcsname{Cost of invariance}
\expandafter\def\csname suppname@sec:hcacls\endcsname{Age classification}
\expandafter\def\csname suppname@sec:rankabl\endcsname{Rank weight ablation}
\expandafter\def\csname suppname@sec:nsens\endcsname{Sensitivity to the retained coefficient count}
\expandafter\def\csname suppname@sec:dc\endcsname{Digit classification}
\expandafter\def\csname suppname@sec:faces\endcsname{Face identification}
\expandafter\def\csname suppname@sec:cifar\endcsname{Natural image classification}
\expandafter\def\csname suppname@sec:modelnet\endcsname{Shape classification on a mesh}
\expandafter\def\csname suppname@sec:simdesign\endcsname{Simulated experiments}
\expandafter\def\csname suppname@sec:pa\endcsname{Pattern classification}
\expandafter\def\csname suppname@sec:oc\endcsname{Object classification}
\expandafter\def\csname suppname@sec:ixipaired\endcsname{Paired comparisons on sex classification}
\expandafter\def\csname suppname@sec:ds\endcsname{Distribution shift}
\expandafter\def\csname suppname@sec:medmnist\endcsname{MedMNIST3D classification}
\expandafter\def\csname suppname@sec:medmnist-protocol\endcsname{MedMNIST3D selection protocol}
\expandafter\def\csname suppname@sec:floatdetail\endcsname{Float construction}
\expandafter\def\csname suppname@sec:protocol\endcsname{Training and statistical protocol}
\expandafter\def\csname suppname@sec:basisfam\endcsname{Basis families}
\expandafter\def\csname suppname@sec:dbscap\endcsname{Neurostimulation generalization}
\expandafter\def\csname suppnum@sec:archcmp\endcsname{A.1}
\expandafter\def\csname suppnum@sec:ca\endcsname{A.2}
\expandafter\def\csname suppnum@sec:encform\endcsname{A.3}
\expandafter\def\csname suppnum@sec:basisprops\endcsname{A.4}
\expandafter\def\csname suppnum@sec:lra\endcsname{A.4.1}
\expandafter\def\csname suppnum@sec:ti\endcsname{A.4.2}
\expandafter\def\csname suppnum@sec:tp\endcsname{A.4.3}
\expandafter\def\csname suppnum@sec:dctklt\endcsname{A.4.4}
\expandafter\def\csname suppnum@sec:corr\endcsname{A.4.5}
\expandafter\def\csname suppnum@sec:magcost\endcsname{A.5}
\expandafter\def\csname suppnum@sec:hcacls\endcsname{A.6}
\expandafter\def\csname suppnum@sec:rankabl\endcsname{A.7}
\expandafter\def\csname suppnum@sec:nsens\endcsname{A.8}
\expandafter\def\csname suppnum@sec:dc\endcsname{A.9}
\expandafter\def\csname suppnum@sec:faces\endcsname{A.10}
\expandafter\def\csname suppnum@sec:cifar\endcsname{A.11}
\expandafter\def\csname suppnum@sec:modelnet\endcsname{A.12}
\expandafter\def\csname suppnum@sec:simdesign\endcsname{A.13}
\expandafter\def\csname suppnum@sec:pa\endcsname{A.13.1}
\expandafter\def\csname suppnum@sec:oc\endcsname{A.13.2}
\expandafter\def\csname suppnum@sec:ixipaired\endcsname{A.14}
\expandafter\def\csname suppnum@sec:ds\endcsname{A.15}
\expandafter\def\csname suppnum@sec:medmnist\endcsname{A.16}
\expandafter\def\csname suppnum@sec:medmnist-protocol\endcsname{A.17}
\expandafter\def\csname suppnum@sec:floatdetail\endcsname{A.18}
\expandafter\def\csname suppnum@sec:protocol\endcsname{A.19}
\expandafter\def\csname suppnum@sec:basisfam\endcsname{A.20}
\expandafter\def\csname suppnum@sec:dbscap\endcsname{A.21}
\expandafter\def\csname suppanchor@sec:archcmp\endcsname{subsection.A.1}
\expandafter\def\csname suppanchor@sec:ca\endcsname{subsection.A.2}
\expandafter\def\csname suppanchor@sec:encform\endcsname{subsection.A.3}
\expandafter\def\csname suppanchor@sec:basisprops\endcsname{subsection.A.4}
\expandafter\def\csname suppanchor@sec:lra\endcsname{subsubsection.A.4.1}
\expandafter\def\csname suppanchor@sec:ti\endcsname{subsubsection.A.4.2}
\expandafter\def\csname suppanchor@sec:tp\endcsname{subsubsection.A.4.3}
\expandafter\def\csname suppanchor@sec:dctklt\endcsname{subsubsection.A.4.4}
\expandafter\def\csname suppanchor@sec:corr\endcsname{subsubsection.A.4.5}
\expandafter\def\csname suppanchor@sec:magcost\endcsname{subsection.A.5}
\expandafter\def\csname suppanchor@sec:hcacls\endcsname{subsection.A.6}
\expandafter\def\csname suppanchor@sec:rankabl\endcsname{subsection.A.7}
\expandafter\def\csname suppanchor@sec:nsens\endcsname{subsection.A.8}
\expandafter\def\csname suppanchor@sec:dc\endcsname{subsection.A.9}
\expandafter\def\csname suppanchor@sec:faces\endcsname{subsection.A.10}
\expandafter\def\csname suppanchor@sec:cifar\endcsname{subsection.A.11}
\expandafter\def\csname suppanchor@sec:modelnet\endcsname{subsection.A.12}
\expandafter\def\csname suppanchor@sec:simdesign\endcsname{subsection.A.13}
\expandafter\def\csname suppanchor@sec:pa\endcsname{subsubsection.A.13.1}
\expandafter\def\csname suppanchor@sec:oc\endcsname{subsubsection.A.13.2}
\expandafter\def\csname suppanchor@sec:ixipaired\endcsname{subsection.A.14}
\expandafter\def\csname suppanchor@sec:ds\endcsname{subsection.A.15}
\expandafter\def\csname suppanchor@sec:medmnist\endcsname{subsection.A.16}
\expandafter\def\csname suppanchor@sec:medmnist-protocol\endcsname{subsection.A.17}
\expandafter\def\csname suppanchor@sec:floatdetail\endcsname{subsection.A.18}
\expandafter\def\csname suppanchor@sec:protocol\endcsname{subsection.A.19}
\expandafter\def\csname suppanchor@sec:basisfam\endcsname{subsection.A.20}
\expandafter\def\csname suppanchor@sec:dbscap\endcsname{subsection.A.21}

%% file: sec/0_abstract.tex
\begin{abstract}
  Vision transformers tokenize an image into patches, which couples the token count to resolution and requires the embedding to be learned from data. We propose a spectral vision transformer architecture for efficient tokenization in the limited-data regime, with an emphasis on medical imaging. Rather than using spatial patches, we tokenize into coefficients of a fixed basis. This approach defines each token as one spectral mode, and the sequence length is chosen rather than implied by resolution, reducing the cost of attention. 
  Our proposed spectral ViT shows superior or comparable performance where data is scarce in public medical imaging benchmarks, natural images, object shapes, simulations, and protected clinical data. We release our code at: \url{https://github.com/agr78/SpectralViT}.
\end{abstract}

%% file: sec/1_intro.tex
\section{Introduction}
\label{sec:introduction}

Vision transformers (ViTs) have proven effective in medical imaging~\cite{zhou2021review,barraganmontero2021review}, capturing global context in segmentation~\cite{ma2024segment}, disease detection~\cite{kim2020mamm,nam2019xr}, and radiological reporting~\cite{bannur2024maira2}. Yet, they require hundreds of thousands of training images, preventing widespread clinical application. These demands may be addressed with transfer learning from unsupervised tasks~\cite{miao2022ssvit} or token-based distillation~\cite{touvron2021deit}. However, this requires massive datasets and pretrained or companion models~\cite{li2022cnnguidedvit} to facilitate ViT training on small datasets. When pretrained on public data, such models are vulnerable to distribution shift~\cite{azizi2022demi,bendavid2010theory}. This presents regulatory challenges~\cite{ranisch2025exp} and degrades performance~\cite{wang2023contrastive,roberts2025prlxgan}, impeding adoption~\cite{piffer2024small}. The lack of large, curated datasets has hindered clinical translation of ViTs to predict outcomes of invasive surgical procedures~\cite{maierhein2022sds}, aggressive conditions~\cite{zhao2022glioma}, and rare diseases~\cite{banerjee2023ml}. To enable the use of ViTs in scarce data, we propose a novel ``spectral'' ViT exploiting modes identified by the selected basis decomposition, illustrated in Figure~\ref{fig:schematic-diagram}. Our contributions include: 
\begin{itemize}
  \item Efficient spectral tokenization that parameterizes task-relevant feature extraction from limited data.
  \item Reduced algorithmic complexity, decoupling the token sequence length and cost of attention from image size.
  \item Robust classification and regression on both subsampled and small datasets against a variety of models.
  \item Specification of theoretical properties guiding the choice of basis.
  \item Extensions that relax the correspondence precondition: overlapping transforms, and fixed nonlinear tokenization.
\end{itemize}
\subsection{Related work}
\label{sec:background}
Spectral transforms have been applied to model inputs~\cite{yao2022wavevit,esteves2025sit}, used to order and embed spatial patches~\cite{wen2025semanticist}, and adopted as fixed-basis tokenizers for time series~\cite{yi2023frequencydomain,masserano2024wavetoken} and spectra~\cite{shen2025wavelength}. Spectral tokens have also supplied multi-modal attention~\cite{wu2025frequencydomainfusion}, autoregression~\cite{huang2025spectralar}, self-regularization~\cite{xiang2025denoisingvit} and information density~\cite{chen2022infodensityvit,chen2022smalldatavit}. Frequency analysis inside the transformer blocks can replace self-attention with spectral filtering~\cite{rao2021gfnet}, interleave spectral layers with attention~\cite{patro2023spectformer}, mix tokens by scattering and spectral gating~\cite{patro2023svt}, gate a spectral branch against window attention~\cite{sun2026spectvit} or modulate attention and fusion in volumetric segmentation~\cite{yang2026feformer}. Hierarchical attention has been used to efficiently embed patches in datasets on the order of tens of thousands of images~\cite{zhang2025dermvit}. Yet these methods all retain expensive patch tokenization. Also related to our method is a natural-image tokenization under a fixed coefficient ordering to enable lossy inference efficiency~\cite{zhou2026frequencyformer} rather than statistical hierarchy. We assert a fixed basis for parameter-free tokenization, derive the embedding hierarchy from the coefficient ordering, and treat the coefficients themselves as tokens. Our approach enables self-attention to model nonlinear dependencies between components, extending modeling capability beyond linear decomposition. Our proposed spectral ViT can be trained from scratch on small sample sizes where pretrained models are unavailable or degraded by distribution shift.

%% file: sec/2_methods.tex
\section{Methods}
\label{sec:methods}
\subsection{Spectral ViT architecture}
To reduce the burden associated with fitting models to small datasets with images $I$ of dimensions $\mathcal{H} \times \mathcal{W}$, the proposed model operates in the span of a \ed{fixed, ordered basis $\{\mathbf{w}_i\}$ rather than on spatial patches. Tokens are the coefficients $s_i$ of the projection of an image $\mathbf{v}$ onto that basis, and the coefficient index forms a rank-ordered sequence resulting in a hierarchical embedding in addition to a spectral positional embedding $\mathbf{E}_{pos}$. Any basis whose coefficients carry an inherent ordering may be used. As the rank weights are trained rather than held at their initialization~(\suppnumref{sec:rankabl}), ordering supplies a prior on the weighting rather than a constraint. Principal component analysis (PCA) is the chosen example below and used in several experiments, expanded in \secname{sec:basegen} and~\suppnumref{sec:basisfam}.} Figure~\ref{fig:schematic-diagram} shows on a sample from the MNIST dataset to illustrate the spectral ViT. \suppnumref{sec:archcmp} summarizes the differences between spatial and spectral ViTs. Generally, the learned linear projection $E$ of a spatial ViT is replaced by an ordered basis, whose coefficients are the tokens. The rank of these spectral coefficients may initialize a hierarchical embedding complementing positional embeddings. Assuming images may be represented in a spectral domain as a \hyperref[sec:precond]{precondition} reduces the number of parameters, as opposed to learning embeddings directly from the image. 
\subsubsection{Spectral tokenization and embedding}
Given a flattened image vector $\mathbf{v} \in \mathbb{R}^{1 \times m}$, tokens are defined by projection onto the $n$ spectral components $\{\mathbf{w}_i\}_{i=1}^{n} \subset \mathbf{W}_{PCA}$, rather than a learned linear projection $\mathbf{E}$ onto $d_e > 1$ dimensional latent space, as in spatial ViTs.
\begin{fequation}\label{eq1}
    S = \{s_1, s_2, \dots, s_n\}, \quad s_i = \mathbf{v} \cdot \mathbf{w}_i = \sum_{j=1}^{m} v_j w_{i,j}
\end{fequation}

Where $w_{i,j}$ is the $i^{th}$ basis vector value at position $j$ weighting raw image $\mathbf{v}$ at position $j$. Each scalar coefficient $s_i$ is mapped to an embedding $\mathbf{h}_i \in \mathbb{R}^{1\times d_e}$ via component projection $\phi_i$. The projection $\phi_i$ consists of weights $\mathbf{W}_\phi \in \mathbb{R}^{n \times d_e}$ and (optional) bias $\mathbf{b}_{\phi,i} \in \mathbb{R}^{1 \times d_e}$. The component index $i$ ordering of the PCA basis is used to initialize a rank weight~\cite{jolliffe2002principal}, $\omega_i = \frac{1}{i}$, which complements spectral positional embeddings of length $d_e$, $\{\mathbf{e}_{\mathrm{pos}}\}_{i=1}^n \subset \mathbf{E}_{pos}$ and places a decreasing prior on components by rank.
\begin{fequation}\label{eq2}
    \begin{split}
    \mathbf{h}_i &= \phi(s_i \cdot \omega_i) + \mathbf{e}_{\mathrm{pos},i} \\
                  &= \left(s_i \cdot \omega_i\right) \mathbf{W}_{\phi,i} + \mathbf{e}_{\mathrm{pos},i}
    \end{split}
\end{fequation}
Here $\mathbf{h}_i \in \mathbb{R}^{1 \times d_e}$ with $d_e \geq 1$. A shared weight $W_{\phi} \in \mathbb{R}^{1 \times d_e}$ may structurally regularize embeddings by limiting spectral components to a common semantic axis defined by $W_{\phi}$ and prevent overfitting. The $n \times d_e$ component-wise form instead ensures the embedding matrix $H$ keeps sufficient rank for attention to learn distinct inter-mode correlations. \ed{The rank weights $\omega_i$ are initialized at $1/i$ by the spectral ordering rather than fixed. We show the effect of initializing and discarding $\omega_i$ in~\suppnumref{sec:rankabl}.}  We emphasize this architecture will admit a general basis with inherent spectral hierarchy and demonstrate additional examples below. 

\begin{figure*}[tb]
  \centering
  \includegraphics[width=\linewidth]{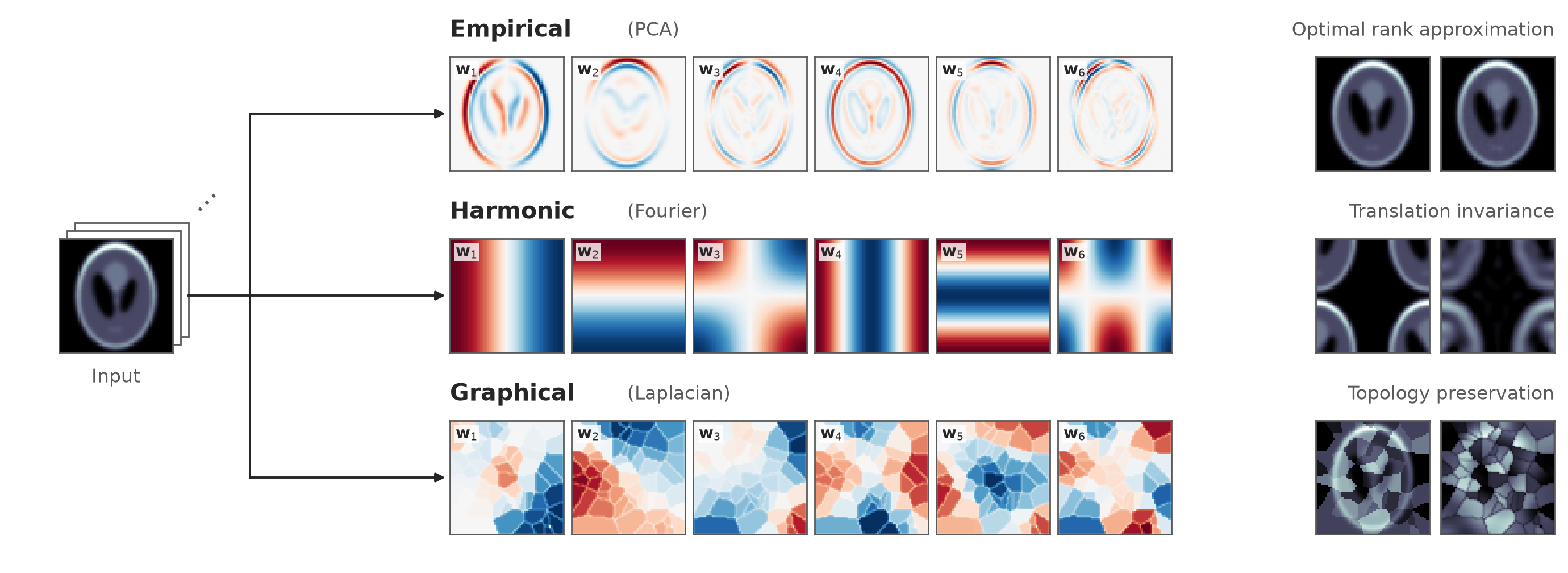}
  \caption{Basis families. Each row contains a basis family, its leading vectors, and an image reconstruction. The empirical basis is fit to a dataset. The harmonic basis is fixed by the image grid with the Fourier transform shown. The graphical basis is fixed by a supplied graph, here the lattice reweighted by a Voronoi partition. Reconstructions under the condition each ordering suits are in~\suppnumref{sec:basisprops}.}
  \label{fig:families}
\end{figure*}

\subsubsection{Bases generalization}\label{sec:basegen}

The framework admits any basis whose coefficients carry an inherent ordering. \ed{Figure~\ref{fig:families} depicts three families of such bases.} An empirical basis is the whitened PCA eigenbasis~(Figure~\ref{fig:families}, top row), whose tokens are the coefficients of Equation~\ref{eq1} ordered by residual variance, attaining the minimum least-squares reconstruction error at any rank~(\suppnumref{sec:lra}). \ed{A harmonic basis~(middle row) is fixed analytically and needs no estimate from the training split as the cosine transform is ordered by frequency~(\suppnumref{sec:dctklt})}. The magnitudes of complex Fourier coefficients are ordered by radial frequency and discard the phase, making the sequence invariant to translation~(\suppnumref{sec:ti}). \ed{A graphical basis~(bottom row) projects onto the eigenvectors of a graph Laplacian, ordered by eigenvalue, so truncation preserves neighborhood structure rather than smoothness on the grid~(\suppnumref{sec:tp}).} Each basis supplies a different notion of a low-rank token. The constructions are given in~\suppnumref{sec:basisfam}, and the properties and reconstructions from tokens in each basis on the Shepp-Logan phantom in~\suppnumref{sec:basisprops}.

\paragraph{Precondition}\label{sec:precond} A fixed linear basis is indexed by coordinates, so its coefficients are informative across samples only when a given coordinate denotes comparable content~(\suppnumref{sec:corr}). \ed{Two extensions relax this requirement. A token need not be a projection - an operation that is nonlinear in the coefficients, such as a magnitude or a topological summary, keeps the fixed tokenization and the rank ordering while suppressing nuisance transformations~(\suppnumref{sec:medmnist}). Overlapping discrete cosine transforms maintain linearity and are indexed by position, so a shared projection is readmitted~(\suppnumref{sec:cifar}).}

\subsubsection{Spectral attention and learned interactions}
The sequence of spectral embeddings $H \in \mathbb{R}^{n \times d_e}$ undergoes self-attention in the same manner as spatial embeddings, following the matrix definition defined by the dot-product between queries $Q$, keys $K$, scaled by the root of key dimension $\sqrt{d_k}$. The softmax is then applied to weight values $V$ to compute $\text{attention}(Q, K, V) = \text{softmax}\left( \frac{QK^T}{\sqrt{d_k}} \right)V$. \ed{Here $Q$, $K$ and $V$ are linear projections of the spectral embeddings themselves, so $V$ carries mode content, as it carries patch content in a spatial ViT, rather than indexing a vocabulary as in language models.}
The spectral token $\mathbf{h}_i$ serves as the input to the attention module. The relationship between spectral component $i$ and component $j$ is defined by the weight $\alpha_{i,j} = \left(e^{\left( \frac{(h_i \mathbf{W}_Q)(h_{j} \mathbf{W}_K)^T}{\sqrt{d_k}} \right)}\right) \left(\sum_{l=1}^n e^{\left( \frac{(h_i \mathbf{W}_Q)(h_l \mathbf{W}_K)^T}{\sqrt{d_k}} \right)}\right)^{-1}$
The contextualized spectral token $\mathbf{h}'_i$ is calculated as the interaction of spectral modes
\begin{fequation}\label{eq5}
    \mathbf{h}'_i = \sum_{k=1}^n \alpha_{i,k} (h_{k} \mathbf{W}_V)
\end{fequation}

Rather than combining these coefficients using fixed weights (as in PCA~\cite{baldi1989neural}) or globally learned weights (as in nonlinear PCA~\cite{kramer1991nonlinear,hsieh2001nonlinear}), we model task-relevant interactions between spectral component embeddings with attention. The encoder is written in the general form by substituting Equation~\ref{eq1} and Equation~\ref{eq2} into Equation~\ref{eq5}, omitting $\mathbf{e}_{\mathrm{pos},i}$ for brevity and derived in full in~\suppnumref{sec:encform}, $f(\mathbf{v}) = g\left( \left\{ \sum_{k=1}^{n} \alpha_{ik}(\mathbf{v}) \, \phi \left(s_{k}, \omega_{k}\right) \mathbf{W}_V \right\}_{i=1}^{n} \right)$
\ed{$g$ receives $n$ tokens, each a sum over $k$. Here $\alpha_{ik}(\mathbf{v})$ are learned interaction weights computed via self-attention between the token at rank $i$ and the token at rank $k$, and $g$ denotes the nonlinear transformation implemented by the transformer layers and prediction head. Learning $\alpha_{ik}$ against the training objective makes the interactions supervised rather than reconstructive. As $\alpha_{ik}(\mathbf{v})$ depends on the input $\mathbf{v}$, the weights are instance-specific rather than globally fixed, which is what distinguishes the model from a linear decomposition or an applied nonlinearity.}

\subsubsection{Complexity}\label{sec:ca-body}
\ed{Tokenization decouples the sequence length from the image resolution. Projecting $\mathbf{v} \in \mathbb{R}^m$ onto $n$ fixed basis vectors costs $\mathcal{O}(nm)$ once, reducing to $\mathcal{O}(m\log m)$ for the bases that admit a fast transform, after which every layer is $\mathcal{O}(nd_e^2 + n^2 d_e)$ in $n$ alone. A spatial ViT on a volume of side $\mathcal{V}$ and patch $\wp$ instead demands $N_P = (\mathcal{V}/\wp)^3$ tokens. Consequently, its per-layer attention is $\mathcal{O}((\mathcal{V}/\wp)^6)$ and grows with resolution at fixed patch size. Our $n$ is set directly, whereas a patch size chosen to fix the token count also fixes what each token spans. The derivation can be found in~\suppnumref{sec:ca}.}

%% file: sec/3_experiments.tex
\section{Experiments}
\label{sec:experiments}
We compare the proposed spectral ViT to a variety of architectures on ten simulated, public and clinical datasets, spanning classification and regression: simulated patterns and objects~(\suppnumref{sec:pa} and~\suppnumref{sec:oc}), MNIST~(\suppnumref{sec:dc}), CIFAR-10 and CIFAR-100~(\suppnumref{sec:cifar}), LFW~(\suppnumref{sec:faces}), ModelNet40~(\suppnumref{sec:modelnet}), IXI~(\secname{sec:sc} and~\suppnumref{sec:ds}), HCP-Aging~(\secname{sec:ba} and~\suppnumref{sec:hcacls}), the six MedMNIST3D cohorts~(\suppnumref{sec:medmnist}) and two neurostimulation cohorts~(\secname{sec:nscc}). \ed{Hardware, optimizers, training budgets and statistical tests are given in~\suppnumref{sec:protocol}, and an augmentation control for the spatial baseline in~\suppnumref{sec:dc}. In every experiment the comparison is paired, with every model scored on the same partition, and capacity is selected inside each training split. 
On the small external cohort of~\secname{sec:nscc} we use permutation tests for significance. The simulated experiments of~\suppnumref{sec:pa} and~\suppnumref{sec:oc} are summarized by the crossover sample size and accuracy respectively, each over $100$ seeds.}

\subsection{Sex classification}\label{sec:sc}
\ed{We compare the proposed spectral ViT with a standard spatial ViT using the Information eXtraction from Images (IXI) public dataset~\cite{ixi_dataset} in the $T_1$-weighted imaging sex classification task.} The IXI dataset, publicly available from \url{brain-development.org}, consists of $566$ acquisitions labeled by sex with fractions $0.45$ male, $0.55$ female across 3 sites. We included two further baselines, the shifted window (Swin) ViT~\cite{liu2021swin} and the U-net with attention~\cite{oktay2018unetatten}, a convolutional encoder of three layers with an attention gate, globally average pooled into one linear layer and trained from scratch as with the other models. Labeled images were center cropped to a cube of dimension $96$ and split into $5$ folds. \ed{The folds are stratified on sex, since a small test fold could otherwise hold a single class, and each partition is shared by every model. Each model has exactly one hyperparameter selected by cross-validation, its capacity. The remaining architecture settings are the released implementation's defaults, unchanged at every sample size. For the spectral ViT these were embedding dimension ($16$), feedforward dimension ($32$), number of heads ($2$) and number of layers ($4$); for the spatial ViT, embedding dimension ($128$), feedforward dimension ($256$), number of heads ($4$) and number of layers ($2$); and for the Swin ViT a window size of $6$, which divides the token grid. The Swin ViT is included as a capacity-matched spatial baseline, at $11{,}470$ weights against the spectral ViT's $11{,}374$. Every model is then selected inside each outer fold by $3$-fold cross-validation on that fold's training split alone, under the one-standard-error rule. 
Grids consisted of the number of retained components for the spectral ViT ($16$, $32$, $64$, $128$, from a whitened PCA refit on each training split, the basis attaining the minimum least-squares reconstruction error at any rank~(\suppnumref{sec:lra})), patch size for the spatial ViT ($8$, $12$, $16$), embedding dimension for the Swin ViT ($16$, $32$, $64$) and base channel width for the U-net ($4$, $8$, $16$). A sample covariance has rank at most one below its training split, so the component grid is truncated at each size to the counts the inner split can estimate. At $n=25$ this leaves none of the four, and the count falls back to the bound itself, $12$; at $n=50$ a single candidate survives, and the full grid only at $n=566$. Capacity is therefore bounded by the inner split rather than the training split, so the model is selected from at most two-thirds of the rank its training data supports. We additionally fit two standard models on the same per-fold components, so that the contribution of attention over spectral tokens can be separated from that of the spectral representation itself: logistic regression and a multilayer perceptron. Each was selected jointly over the component count and its own capacity, weight decay ($0.5$, $0.05$, $0.005$) and hidden width ($2$, $6$, $16$) respectively, under the same protocol, optimizer and schedule as the networks. Candidates were compared at $200$ epochs and the selected model refit for $500$, in both cases with a binary cross-entropy loss at an effective batch size of $8$ using AdamW with a one-cycle schedule peaking at learning rate $1\times 10^{-4}$, for a fixed budget with no checkpoint selection.} All models were tested on the withheld fold, noting mean accuracy and AUC across folds. We report these results in Table~\ref{table:ixi}.

\begin{table*}[tb]
  \caption{Sex classification on IXI by training set size: mean fold AUC $\pm$ standard deviation over draws, ten at $n=25$ and $n=50$ and five above, each an independent subsample and, at the full cohort, an independent fold partition. Bold marks the best in each column. An asterisk marks a value significantly above chance over that size's draws (one-sided, $p<0.05$); the five unmarked values are not distinguishable from $0.5$. Parameter counts are fold means at $n=566$, where every model's capacity is selected inside its training split; the selection protocol is in~\suppnumref{sec:floatdetail}.}
  \label{table:ixi}
  \centering
  \begin{small}
  \begin{tabular*}{\textwidth}{@{\extracolsep{\fill}}lrccccc@{}}
    \toprule
    Model & Param. & $n{=}25$ & $n{=}50$ & $n{=}100$ & $n{=}200$ & $n{=}566$ \\
    \midrule
    Logistic regression & $\bf{96}$ & $0.550 \pm 0.105$ & $0.577 \pm 0.058$\rlap{$^{*}$} & $0.694 \pm 0.035$\rlap{$^{*}$} & $\bf{0.799 \pm 0.021}$\rlap{$^{*}$} & $0.880 \pm 0.006$\rlap{$^{*}$}  \\
    Multilayer perceptron & 1,153 & $0.563 \pm 0.159$ & $0.583 \pm 0.063$\rlap{$^{*}$} & $\bf{0.749 \pm 0.068}$\rlap{$^{*}$} & $0.798 \pm 0.020$\rlap{$^{*}$} & $0.885 \pm 0.006$\rlap{$^{*}$}  \\
    U-net with attention & 74,052 & $0.540 \pm 0.151$ & $0.621 \pm 0.053$\rlap{$^{*}$} & $0.670 \pm 0.073$\rlap{$^{*}$} & $0.720 \pm 0.039$\rlap{$^{*}$} & $\bf{0.898 \pm 0.007}$\rlap{$^{*}$}  \\
    Swin ViT & 11,470 & $0.593 \pm 0.106$\rlap{$^{*}$} & $0.583 \pm 0.077$\rlap{$^{*}$} & $0.554 \pm 0.083$ & $0.593 \pm 0.033$\rlap{$^{*}$} & $0.640 \pm 0.027$\rlap{$^{*}$}  \\
    Spatial ViT & 616,101 & $0.508 \pm 0.172$ & $0.619 \pm 0.078$\rlap{$^{*}$} & $0.672 \pm 0.051$\rlap{$^{*}$} & $0.697 \pm 0.033$\rlap{$^{*}$} & $0.734 \pm 0.033$\rlap{$^{*}$}  \\
    \bf{Spectral ViT (proposed)} & 11,374 & $\bf{0.697 \pm 0.149}$\rlap{$^{*}$} & $\bf{0.685 \pm 0.052}$\rlap{$^{*}$} & $0.748 \pm 0.048$\rlap{$^{*}$} & $0.776 \pm 0.026$\rlap{$^{*}$} & $0.845 \pm 0.010$\rlap{$^{*}$}  \\
    \bottomrule
  \end{tabular*}
  \end{small}
\end{table*}

\subsection{Age estimation}\label{sec:ba}
\ed{We evaluate data efficiency on the HCP-Aging release~\cite{bookheimer2019hcpaging}, obtained from the NIMH Data Archive under a Data Use Certification, using the $T_1$-weighted scan of every subject with a reported age ($1{,}177$ subjects, ages $36$--$90$, mean $60.6$). Volumes are center cropped to a cube of dimension $96$ and intensity standardized per scan. Age is predicted as a continuous target and depends on finer-scale structure than sex and is correspondingly more exposed to residual misalignment in native-space crops. We therefore select a decomposition that is blockwise rather than volume-spanning. Tokens are the $K$ lowest frequencies of each cell of a $4\times4\times4$ block decomposition, one per conjugate pair and log compressed, giving $4^3 \times 4^3 = 4096$ coefficients~(\suppnumref{sec:dctklt} and~\suppnumref{sec:nsens}). For the spectral ViT these are ordered by frequency, which approximates the variance ordering within a block~(\suppnumref{sec:dctklt}), and standardized per feature using the training split alone, matching the whitening applied to the PCA basis elsewhere. Subjects are subsampled at $n \in \{25, 50, 100, 200, 1000\}$, stratified on a median split of age ($61$ years, $576$ against $601$) so that the age distribution is preserved at every size, with $5$-fold cross-validation and independent subsamples per size, ten at $n=25$ and $n=50$ and five above. Every model's capacity is selected inside each outer fold by $3$-fold cross-validation on that fold's training split alone, over the number of embedding dimensions for the transformers, hidden width for the perceptron, patch size for the spatial ViT, base channel width for the U-net and weight decay for logistic regression. The one-standard-error rule selects the smallest capacity statistically indistinguishable from the best using only training data. Models were trained with an $L_1$ loss at learning rate $1\times10^{-3}$. We report mean absolute error in years, pooled over withheld folds, in Table~\ref{table:hca-reg}.}

\begin{table*}[tb]
  \caption{Age estimation on HCP-Aging. Mean absolute error in years, pooled over withheld folds, $\pm$ standard deviation over independent subsamples, of which there are $10$ at $n=25$ and $n=50$ and $5$ at the larger sizes. Lower is better, and bold marks the best value in each column. Every model improves on the mean estimate at every sample size ($p \leq 0.005$, paired over draws), so no cell is a restatement of the cohort mean. Capacity selection and the parameter counts are described in the supplementary material.}
  \label{table:hca-reg}
  \centering
  \begin{small}
  \begin{tabular*}{\textwidth}{@{\extracolsep{\fill}}lrccccc@{}}
    \toprule
    Model & Parameters & $n=25$ & $n=50$ & $n=100$ & $n=200$ & $n=1000$ \\
    \midrule
    Mean estimate                  & -- & $12.96 \pm 0.67$ & $13.49 \pm 0.88$ & $13.79 \pm 1.11$ & $14.09 \pm 1.04$ & $13.84 \pm 0.06$  \\
    \midrule
    Logistic regression            & $\bf{4,097}$ & $11.07 \pm 1.22$ & $8.43 \pm 0.75$ & $6.51 \pm 0.24$ & $5.92 \pm 0.30$ & $5.20 \pm 0.11$  \\
    Multilayer perceptron          & 43,768 & $9.05 \pm 0.96$ & $7.96 \pm 0.89$ & $6.64 \pm 0.36$ & $5.99 \pm 0.32$ & $\bf{4.92 \pm 0.31}$  \\
    U-net with attention           & 18,387 & $9.04 \pm 1.04$ & $7.95 \pm 0.73$ & $6.63 \pm 0.23$ & $6.11 \pm 0.28$ & $5.20 \pm 0.10$  \\
    Swin ViT                       & 13,682 & $10.68 \pm 1.69$ & $10.41 \pm 1.04$ & $9.88 \pm 0.93$ & $9.04 \pm 1.17$ & $7.40 \pm 0.17$  \\
    Spatial ViT                    & 65,554 & $10.86 \pm 1.92$ & $9.39 \pm 0.84$ & $7.95 \pm 1.38$ & $6.30 \pm 0.25$ & $5.05 \pm 0.23$  \\
    \bf{Spectral ViT (proposed)}   & 10,149 & $\bf{7.64 \pm 1.14}$ & $\bf{7.05 \pm 0.84}$ & $\bf{6.20 \pm 0.08}$ & $\bf{5.88 \pm 0.22}$ & $5.23 \pm 0.16$  \\
    \bottomrule
  \end{tabular*}
  \end{small}
\end{table*}

\subsection{Neurostimulation candidate classification}\label{sec:nscc}
\begin{figure*}[tb]
  \centering
  \includegraphics[width=\linewidth]{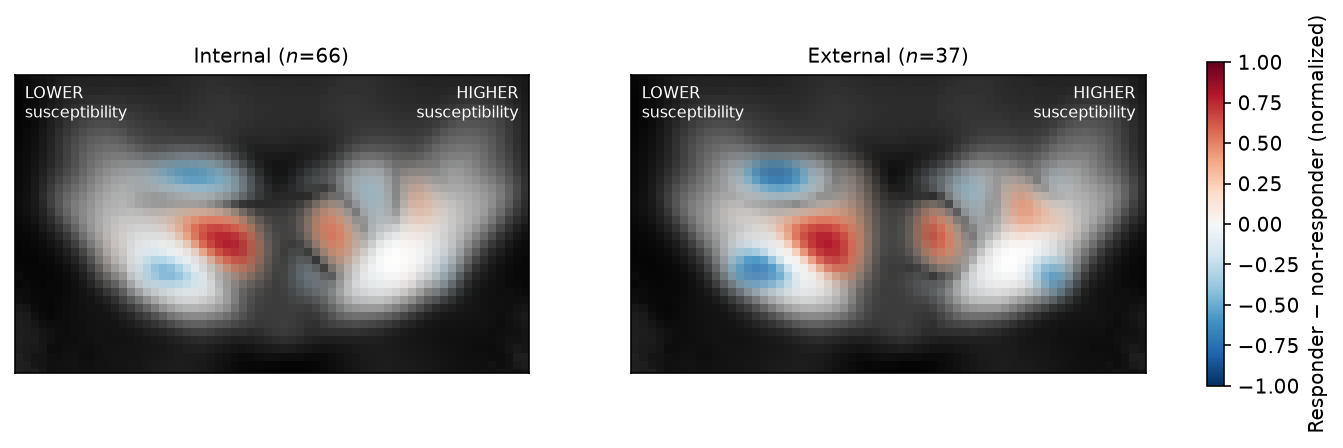}
  \caption{\ed{Spectral modes that separate neurostimulation responders consistently at both sites, shown as a responder minus non-responder difference in the image domain. The panels are the internal cohort ($n=66$) and the external cohort ($n=37$), each crop reconstructed from the retained modes.
  The two nuclei are grouped by mean susceptibility rather than hemisphere, as is the model's input. 
  The selection rule and the windowing are in the supplementary material.}}
  \label{fig:dgn-diagram}
\end{figure*}

Whole-brain quantitative susceptibility maps (QSM)~\cite{roberts2024msmv} were reconstructed for $66$ patients from an internal dataset and $37$ patients from an external dataset undergoing bilateral STN-DBS~\cite{roberts2026dbs}. The label is defined from the Unified Parkinson’s Disease Rating Scale (UPDRS) motor score off medication: the six-month postsurgical score on stimulation against the presurgical baseline, response being an improvement of at least $30\%$, the conventional threshold in this literature. An atlas and masks containing deep gray nuclei DBS targets were generated. \ed{Each nucleus was cropped to $20 \times 20 \times 16$ voxels about its center of mass, and the two nuclei were ordered by mean susceptibility so that the representation is independent of hemisphere. Analysis is at the patient level.
Responders outnumber non-responders roughly $12:1$ in both cohorts ($61:5$ internally, $34:3$ externally), so balanced accuracy and specificity rather than accuracy are the informative metrics. The $K$ lowest Fourier frequencies of each crop were retained, one per conjugate pair and ordered by frequency, and their magnitudes log compressed, giving $2K$ spectral tokens. The magnitude discards phase so that a token is invariant to the residual translation of the atlas-centered crops~(\suppnumref{sec:ti}). Seven clinical covariates are individually tokenized (age, sex, disease duration, LEDD, presurgical UPDRS off medication ($u_{\mathrm{off}}$) and on medication ($u_{\mathrm{on}}$), and levodopa responsiveness $(u_{\mathrm{off}}-u_{\mathrm{on}})/u_{\mathrm{off}}$). We note that $u_{\mathrm{off}}$ is also the denominator of the response label, following improvement definition convention~\cite{defer1999capsit}. Each model's capacity was selected by internal leave-one-out AUC alone under the one-standard-error rule, over $K \in \{16,32\}$ for the spectral models, the corresponding patch size for the spatial ViT, and base channel width for the U-net. 
Models were trained with a binary cross-entropy loss for $200$ full-batch steps using AdamW with learning rate $3\times 10^{-3}$ and weight decay $0.1$. Internal performance is estimated by leave-one-out cross-validation across all $66$ patients. External predictions come from a single model refit on all internal data. Scores were averaged over $10$ random initializations and thresholded once to maximize internal balanced accuracy.
Because the two cohorts differ in score distribution, its quantile rather than its value is transferred, which requires only comparable non-responder prevalence ($7.6\%$ versus $8.1\%$). 
The operating point is chosen on internal data alone, and the model score distribution from external inference
maps that quantile back to a threshold. 
} 
In Table~\ref{table:dbs}, we compare the proposed spectral ViT to logistic regression and a transformer on the clinical covariates, to logistic regression and a multilayer perceptron on the same spectral features, and to two spatial-domain baselines: a spatial ViT and an attention U-net. Figure~\ref{fig:dgn-diagram} shows the spectral modes that separate responders at both sites. \ed{The neurostimulation cohorts are comparable to those described in~\cite{roberts2026dbs}. A retrospective study approved by the institutional review boards and ethics committees of the relevant institutions, with informed consent obtained and every patient anonymized before analysis.}

\begin{table*}[!tb]
  \caption{Neurostimulation candidate classification on the external cohort, thresholded at an operating point calibrated on internal data alone. Capacity is selected within the internal cohort by the one-standard-error rule
  The spectral ViT leads on external AUC and is the only model whose internal estimate anticipates its external one. 
  The $p$ column is a one-sided permutation test on external AUC.} 
  \label{table:dbs}
  \centering
  \begin{small}
  \begin{tabular*}{\textwidth}{@{\extracolsep{\fill}}llllll@{}}
    \toprule
    Model & Capacity & Balanced accuracy & Specificity & AUC & $p$ \\
    \midrule
    Clinical logistic regression & -- & $0.539$ & $0.667$ & $0.667$ & $0.188$  \\
    Clinical transformer & -- & $0.593$ & $0.333$ & $0.637$ & $0.236$  \\
    Spectral logistic regression & $16$ modes & $0.569$ & $0.667$ & $0.765$ & $0.074$  \\
    Spectral MLP & $16$ modes & $0.657$ & $0.667$ & $0.657$ & $0.203$  \\
    U-net with attention & $12$ channels & $0.529$ & $\bf{1.000}$ & $0.392$ & $0.729$  \\
    Spatial ViT & patch $4$ & $0.529$ & $\bf{1.000}$ & $0.794$ & $0.052$  \\
    \bf{Spectral ViT (proposed)} & $\mathbf{16}$ \bf{modes} & $\bf{0.794}$ & $\bf{1.000}$ & $\bf{0.814}$ & $\bf{0.040}$  \\
    \bottomrule
  \end{tabular*}
  \end{small}
\end{table*}

\begin{figure*}[tb]
  \centering
  \includegraphics[width=\linewidth]{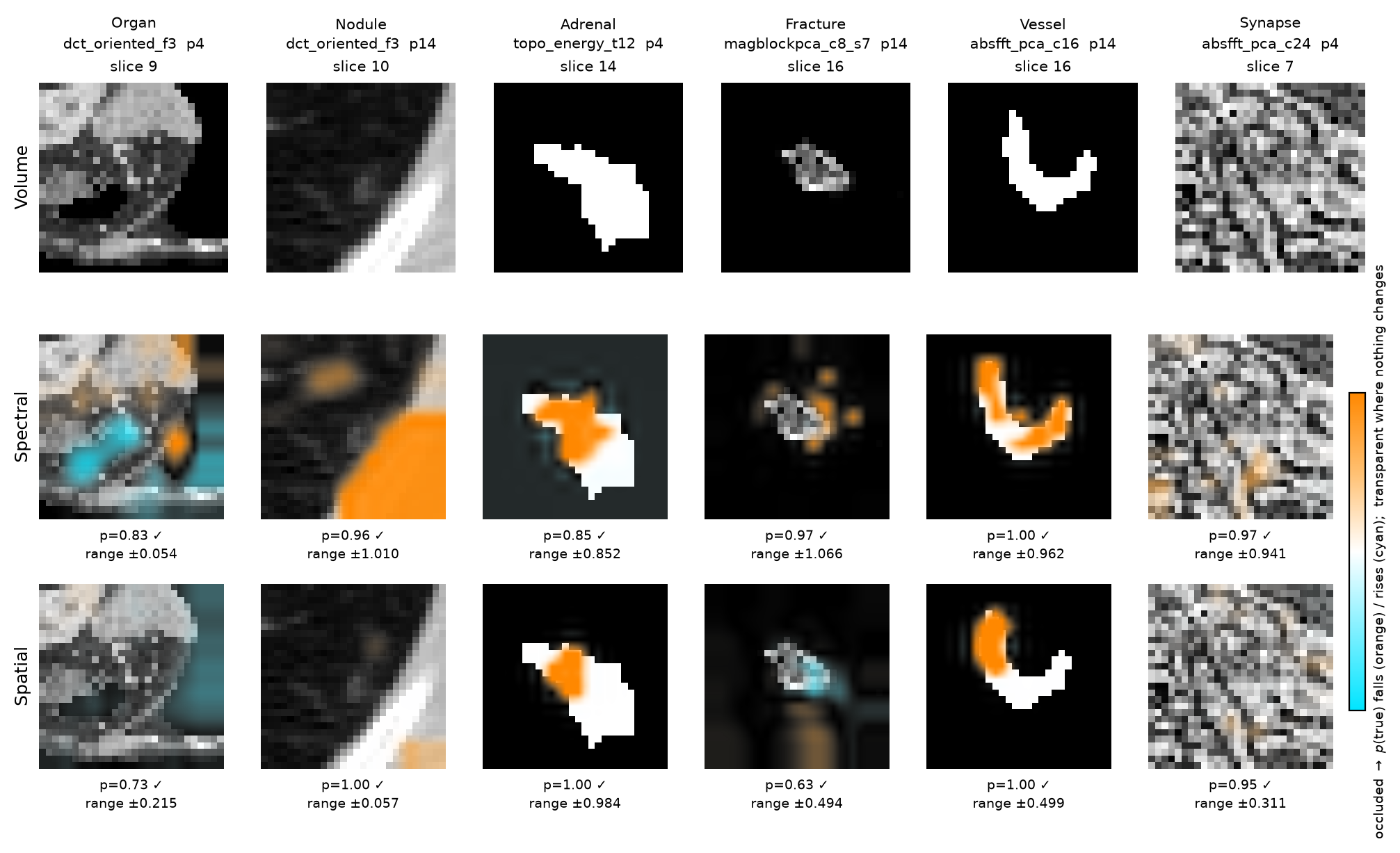}
  \caption{\ed{Occlusion test on one withheld volume per MedMNIST3D cohort at $n=25$. Row~1 is the volume. Rows~2 and~3 overlay occlusion saliency for the spectral and spatial models on that slice.
  Red marks windows whose removal lowers $p(\text{true class})$, blue those that raise it, transparent regions those that change nothing. 
  The maps 
  separate the cohorts into two regimes: where the structure is a sparse mask 
  and a global magnitude basis is barely perturbed, consistent with SynapseMNIST3D, the densest, being the cohort where the spectral model fares worst. Construction is in~\protect\suppnumref{sec:medmnist}.}}
  \label{fig:medmnist-saliency}
\end{figure*}

%% file: sec/4_results.tex
\section{Results}\label{sec:results}

\ed{On~\secname{sec:sc} the spectral ViT leads every baseline at the two smallest sample sizes, as reported in Table~\ref{table:ixi}. At $n=50$ it is ahead in every draw against four of the five baselines; at $n=25$ the margins are wider but so is the spread across draws, and only the comparison against the U-net separates individually
~(\suppnumref{sec:ixipaired}). 
The spectral ViT does this with $11{,}374$ parameters against the standard spatial ViT's $616{,}101$, and the full nested evaluation, inner capacity selection included, takes $91$ minutes against $4.7$ hours on the same hardware.\footnote{Excludes the fixed PCA basis preprocessing, $\mathcal{O}(nm)$.} The ordering inverts as the cohort grows, expected for an efficient tokenization. The multilayer perceptron is comparable at $n=100$, 
both coefficient-input baselines lead from $n=200$, 
and the translationally equivariant U-net with attention overtakes it only at the full cohort, with $6.5$ times the parameters. This cost of truncation is expected where the signal is partly anatomical shape. At most $128$ components describe the volume in our spectral ViT, while an equivariant convolutional decoder retains all of it and is appropriate for unregistered native-space acquisitions.
State of the art approaches on IXI sex classification combine anatomical priors~\cite{shehata2022gnnixi} and additional datasets~\cite{juglan2025ixi} to achieve between $0.90$ and nearly perfect AUC. We show in~\suppnumref{sec:ds} that this performance degrades with noise in the test distribution.}
\ed{Briefly, distribution mismatch is tested on the~\secname{sec:sc} task against transfer learning from UK Biobank, with Rician noise added to the withheld fold at test time only (\suppnumref{sec:ds}). The pretrained model leads on uncorrupted images until SNR $4$, then collapses to chance at SNR $1$ and below, where the spectral ViT keeps AUC above $0.74$~(\suppnumref{sec:ds}).}
\ed{On the~\secname{sec:ba} task the spectral ViT attains the lowest error at every sample size up to $n=200$ and is overtaken at $n=1{,}000$, as reported in Table~\ref{table:hca-reg}. Its advantage over the translation-equivariant U-net is largest in the smallest sample 
and narrows monotonically with the cohort.
The same pattern holds against logistic regression and against the perceptron applied to the same coefficients. 
The attention mechanism advantage over a perceptron on the same coefficients recedes once the sample is large. The spectral tokenization contribution is concentrated in the small-sample regime as expected.
Every model improves upon the mean estimator, 
reflecting learned structure rather than regression toward the mean. 
Replacing the block cosine transform with the magnitude of a block Fourier transform, which discards phase in exchange for translation invariance, 
adds as much as $1.58$ years of error~(\suppnumref{sec:magcost}). Conversely, invariance improves results in~\suppnumref{sec:oc}, where position is a nuisance.}

\ed{The framework extends to operations that are not linear projections of $\mathbf{v}$, reported on all six MedMNIST3D benchmarks~\cite{yang2023medmnist}, each training set subsampled to $n \in \{25,\dots,800\}$~(\suppnumref{sec:medmnist}), where the crossover reproduces.
Pooled over cohorts the spectral model leads by $0.060$ AUC at $n=25$ and the margin decays monotonically with the sample, reaching parity between $n=200$ and $n=400$. Counted per cohort the same shape appears: the spectral model prevails at $n=25$, $4$ of $6$ at $n=50$ and $n=100$, and $3$ of $6$ at $n=200$.
The cohort-wise table and the protocol parameters are given in~\suppnumref{sec:medmnist}. 
Figure~\ref{fig:medmnist-saliency} analyzes one correctly classified test volume per cohort at $n=25$. Attribution is by occlusion rather than gradient so that the two models are comparable.
}

\ed{Finally, on the deep brain stimulation cohorts of~\secname{sec:nscc} the spectral ViT attains the highest external AUC in Table~\ref{table:dbs} ($0.814$, exact permutation $p=0.040$), with the spatial ViT next at $0.794$ ($p=0.052$) and the clinical covariate model at $0.667$ ($p=0.188$). Pairwise margins are unresolved at this sample size (\suppnumref{sec:dbscap}). External rank is not always informative here, because across the capacity grid the internal and external AUCs are uncorrelated ($r=-0.03$, $p=0.92$). The spatial ViT reaches $0.794$ externally from an internal estimate of $0.287$, below chance at every capacity. 
Only the spectral ViT offers an internal estimate correlated to its external one, giving balanced accuracy $0.787$ internally against $0.794$ externally, a gap of $0.007$ where logistic regression on the same coefficients loses $0.185$ and the clinical covariate model $0.182$. A single fit takes under a second. The full 
evaluation over all patients and initializations takes eight minutes.}
\ed{Both spatial-domain baselines discriminate below chance on internal cross-validation ($0.287$ and $0.409$), so selecting their capacity on that criterion is uninformative.
The spectral ViT is the only model correlating internal and external estimates. The capacity grid, the null spread at this sample size, the cross-site consistent modes of Figure~\ref{fig:dgn-diagram} and the per-patient comparison with the clinical model are given in~\suppnumref{sec:dbscap}.}

\ed{Additional benchmark experiments confirm these trends. On LFW the spectral ViT leads the patch tokenizer by $0.047$ top-1 at a single image per identity, rising to $0.146$ at ten, and leads eigenfaces and a multilayer perceptron on the same sixteen components at every size~(\suppnumref{sec:faces}). On CIFAR-10 and CIFAR-100, where correspondence fails, the overlapping block cosine extension exceeds a patch tokenizer run at the same stride by $0.017$, $0.037$ and $0.037$ AUC at $n=25$, $50$ and $200$~(\suppnumref{sec:cifar}). On ModelNet40, where no patch tokenizer exists, the graph Laplacian basis leads a projection learned end to end by $0.071$ and $0.047$ top-1 at $40$ and $80$ shapes and matches it in AUC throughout~(\suppnumref{sec:modelnet}).}

%% file: sec/5_limitations.tex
\section{Discussion}\label{sec:discussion}
\ed{Across the cohorts where spatial correspondence holds, the spectral tokenization leads at the smallest sample sizes and is overtaken as the cohort grows. Our proposed spectral ViT allocates a fixed or label-free basis rather than scarce labeled data to learning the embedding. We do not assert spectral tokenization as a general replacement for a learned patch embedding. The basis is an interpretable modeling decision rather than an implementation detail, since selection depends on the structure of the task~(\suppnumref{sec:basisfam}).} \ed{Our tokenization is a fixed linear map of the voxel coordinates, and no such map other than the sample mean is invariant to translation~(\suppnumref{sec:corr}), so a global fixed basis requires that a given coordinate denote the same content in every sample. Where correspondence is absent, comparable performance to spatial models can be achieved by enforcing invariance with a nonlinearity~(\suppnumref{sec:ti}), or a localized basis~(\suppnumref{sec:cifar}). Both extensions keep the basis fixed and the embedding unestimated: the first operates on magnitudes~(\suppnumref{sec:magcost}), the second on local blocks whose token count follows the stride~(\suppnumref{sec:cifar}). The default global basis is therefore bounded to data in a common coordinate frame.} \ed{The basis is itself a modeling choice carrying its own failure modes, the spectral ViT inheriting noise sensitivity, aliasing and graph sensitivity from the eigenbasis, Fourier and Laplacian constructions respectively.} We introduce bias by fixing the projection rather than learning it. As such, when sufficient data is available, learned embeddings will likely outperform our method. 
\section{Conclusion}
We propose a novel ViT with tokens in the ordered spectral rather than image domain. \ed{Our architecture reduces parameter burden by replacing the learned linear projection with a spectral decomposition and a rank hierarchy initialized from component ordering, at under $2\%$ of the parameters and a third of the training time.
Our method leads at the smallest sample sizes in every experiment where the precondition holds: brain MRI tasks~(\secname{sec:sc},~\secname{sec:ba}), all six MedMNIST3D cohorts~(\suppnumref{sec:medmnist}), MNIST~(\suppnumref{sec:dc}), natural faces~(\suppnumref{sec:faces}), simulated patterns~(\suppnumref{sec:pa}), and shapes on a graph 
~(\suppnumref{sec:modelnet}). We attain the best generalization in~\secname{sec:nscc}. We provide two extensions for the case where the spatial correspondence precondition is not met: a nonlinear tokenization that discards phase~(\suppnumref{sec:ti}), and a block spectral extension, with which the spectral tokenizer leads or matches the patch tokenizer at every sample size on both datasets,
 ~(\suppnumref{sec:cifar}).
As modeling choice, the properties a basis supplies can be matched to the structure of a task.} 
\ed{The tokenization behind our proposed spectral ViT alleviates the data and computational demands that prohibit conventional ViTs in the limited data regimes ubiquitous in medical imaging.}

%% file: sec/X_suppl.tex
\ifsuppinline\clearpage\fi
\appendix
\section{Supplementary material}\label{sec:suppl}

\subsection{Architecture comparison}\label{sec:archcmp}
\ed{Table~\ref{table:comparison} sets the two tokenizations side by side on the properties Figure~\ref{fig:schematic-diagram} illustrates: the domain the tokens live in, the inductive bias each carries, what a token and its position are, what attention relates, and how the image is represented.}

\begin{table*}[t]
  \caption{Architecture comparison}
  \label{table:comparison}
  \centering
  \begin{small}
  \begin{tabular*}{\textwidth}{@{\extracolsep{\fill}}lll@{}}
    \toprule
    Property              & Spatial ViT  & Spectral ViT (proposed) \\
    \midrule
    Domain                & Image space ($\mathbb{R}^{\mathcal{H} \times \mathcal{W}}$)  & Eigenspace ($\mathbb{R}^{n_{\mathrm{components}}})$ \\
    Inductive bias        & Spatial localization                     & Variance hierarchy \\
    Token                 & Image patch                              & Spectral component \\
    Position              & Grid coordinates                         & Spectral coordinates \\
    Relationship          & $\alpha_{i,j} \propto$ spatial proximity & $\alpha_{i,j} \propto$ component interaction \\ 
    Representation        & $I_{\mathrm{spatial}} \approx$ \text{concat}($p_1,...,p_{n_\mathrm{patches}}$) & $I_{\mathrm{spectral}} \approx \sum_{i=1}^{n_\mathrm{components}} s_i \mathbf{w}_i$ \\
    \bottomrule
  \end{tabular*}
  \end{small}
\end{table*}

\subsection{Complexity analysis}\label{sec:ca}
We evaluate the computational complexity of the Spectral ViT in three stages: spectral tokenization ($\text{cost}_{\text{spec}}$), component-wise embedding ($\text{cost}_{\text{embed}}$), and the transformer backbone ($\text{cost}_{\text{trans}}$). Let $m$ remain the total number of voxels ($\mathcal{V}^3$ for a volume of dimension $\mathcal{V}$), $n$ the number of spectral components, $d_e$ the embedding dimension, and $L$ the number of transformer layers. The total cost is formulated as:
\begin{fequation}
    \text{cost}_{\text{total}} = \text{cost}_{\text{spec}} + \text{cost}_{\text{embed}} + L \times \text{cost}_{\text{trans}}
\end{fequation}
The initial stage involves projecting the raw image $\mathbf{v} \in \mathbb{R}^m$ onto a fixed PCA basis to extract $n$ scalar coefficients with matrix-vector multiplication, incurring a single linear cost relative to voxels $m$ prior to training and inference:
\begin{fequation}
    \text{cost}_{\text{spec}} = n \times m = \mathcal{O}(nm)
\end{fequation}
Each scalar coefficient $s_i$ is mapped to the embedding space via a component-wise projection weight $\mathbf{W}_{\phi,i} \in \mathbb{R}^{1 \times d_e}$, which incurs:
\begin{fequation}
    \text{cost}_{\text{embed}} = n \times d_e = \mathcal{O}(nd_e)
\end{fequation}
Within each layer $L$, weight matrices project the embeddings into queries, keys, and values, followed by the self-attention mechanism
\begin{fequation}
    \text{cost}_{\text{trans}} = \underbrace{3(n \cdot d_e^2)}_{\text{Projection}} + \underbrace{2(n^2 \cdot d_e)}_{\text{Attention}} = \mathcal{O}(nd_e^2 + n^2d_e)
\end{fequation}

The spectral ViT efficiently decouples image resolution $m$ from the transformer sequence length $n$. \ed{The spatial ViT~\cite{dosovitskiy2021vit} sequence length is $N_P=\left(\frac{\mathcal{V}}{\wp}\right)^3$, where $\mathcal{V}$ is the dimension of the cubic input volume and $\wp$ is the cubic patch size. Quadratic self-attention incurs a per-layer complexity of $\mathcal{O}(N_P^2)=\mathcal{O}\left(\left(\frac{\mathcal{V}}{\wp}\right)^6\right)$. Both architectures incur an initial voxel-processing cost ($\mathcal{O}(nm)$ for spectral tokenization and $\mathcal{O}(md_e)$ for spatial patch embedding), after which the self-attention complexity of subsequent transformer layers is reduced from $\mathcal{O}\left(\left(\frac{\mathcal{V}}{\wp}\right)^6\right)$ to $\mathcal{O}(n^2)$. The same reduction applies to any fixed basis. Where a fast transform exists, as for the Fourier and cosine bases, the initial projection costs $\mathcal{O}(m\log m)$ rather than $\mathcal{O}(nm)$; a graph Laplacian basis admits no such factorization in general and stays at $\mathcal{O}(nm)$, which is paid once and does not enter the per-layer cost either way. As $n \ll N_P$ in typical volumetric imaging settings, this shift offers a significant reduction in the computational resources required for self-attention for 3D medical imaging.}

\subsection{Encoder form}\label{sec:encform}
\ed{The general form quoted in~\secname{sec:methods} is three substitutions, with $\mathbf{e}_{\mathrm{pos}}$ omitted throughout for brevity. Equation~\ref{eq1} gives the coefficient at rank $k$, Equation~\ref{eq2} embeds it, and we write $\phi(s_k,\omega_k) = (s_k \omega_k)\mathbf{W}_{\phi,k}$ for that embedding:}
\begin{fequation}
    s_k = \mathbf{v}\cdot\mathbf{w}_k, \qquad
    \mathbf{h}_k = \phi(s_k,\omega_k) = (s_k \omega_k)\,\mathbf{W}_{\phi,k}.
\end{fequation}
\ed{Equation~\ref{eq5} contextualizes rank $i$ against every rank $k$, so substituting $\mathbf{h}_k$ gives}
\begin{fequation}
    \mathbf{h}'_i = \sum_{k=1}^{n} \alpha_{ik}(\mathbf{v})\,\bigl(\mathbf{h}_k \mathbf{W}_V\bigr)
                  = \sum_{k=1}^{n} \alpha_{ik}(\mathbf{v})\,\phi(s_k,\omega_k)\,\mathbf{W}_V .
\end{fequation}
\ed{The encoder returns one such token per rank. Collecting all $n$ of them and applying $g$, the remaining layers and the prediction head, gives the quoted form:}
\begin{fequation}
    f(\mathbf{v}) = g\left( \left\{ \sum_{k=1}^{n} \alpha_{ik}(\mathbf{v})\, \phi\left(s_{k},\omega_{k}\right) \mathbf{W}_V \right\}_{i=1}^{n} \right).
\end{fequation}
\ed{Two readings of the notation are worth stating. The braces are a sequence of length $n$, one entry per output rank $i$, not a second summation: the only sum is the inner one over $k$, and it is evaluated once for each $i$. And the attention weight carries the argument $\mathbf{v}$ because $\alpha_{ik}(\mathbf{v}) \propto \exp\bigl((\mathbf{h}_i\mathbf{W}_Q)(\mathbf{h}_k\mathbf{W}_K)^{\!\top}/\sqrt{d_k}\bigr)$ is a function of the embeddings, which are themselves functions of the coefficients. That dependence is the whole of the model's nonlinearity in $\mathbf{v}$: $\mathbf{W}_{\phi,k}\mathbf{W}_V$ is a product of two fixed matrices, so holding $\alpha$ constant in $\mathbf{v}$ collapses $f$ to a linear decomposition followed by $g$. Learning $\alpha$ against the training objective is what separates the model from one.}

\subsection{Spectral basis properties}\label{sec:basisprops}
\begin{figure*}[tb]
  \centering
  \includegraphics[width=\linewidth]{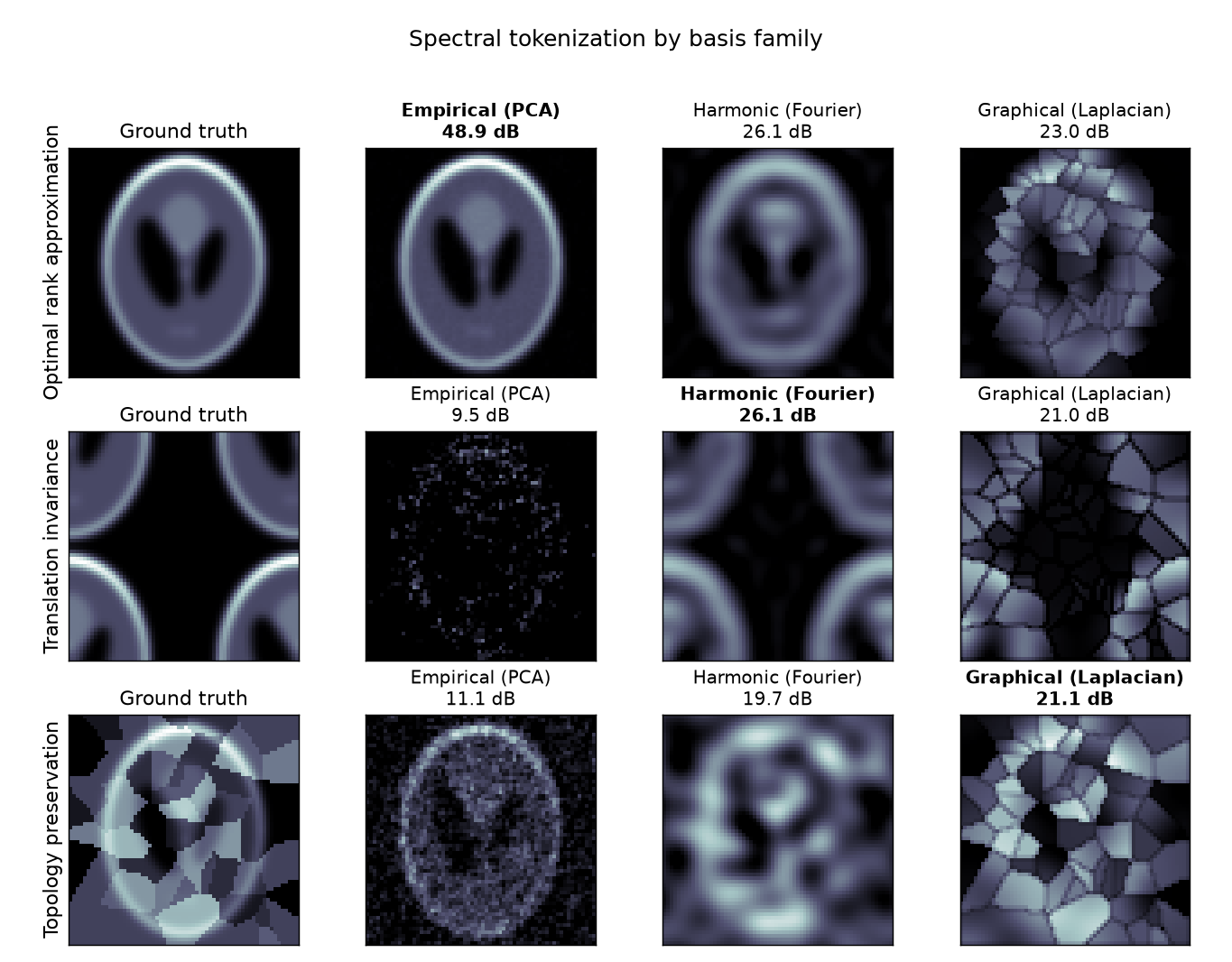}
  \caption{Spectral tokenization and reconstruction by basis family. The columns are the three families and the rows are one property each, constructed so that exactly one family is the right choice for the row, and each family wins its own. Every panel keeps $128$ coefficients of the same noisy image and is scored by PSNR against the clean ground truth. The graphical column diagonalizes the Laplacian of a single fixed Voronoi-weighted pixel lattice shared by every row, and its near-zero cross-region weights make those reconstructions piecewise constant even where the content is not. Bold marks the best value in each row; the construction is in~\protect\secname{sec:tp}.}
  \label{fig:bases}
\end{figure*}
\subsubsection{Rank-$n$ optimality}\label{sec:lra}
Tokenization using the PCA basis $\mathbf{W}_{PCA}$ yields the minimum least-squares reconstruction error. Given a centered image signal $\mathbf{v}$, the spectral coefficients $\mathbf{s}$ provide the best $n$-rank linear approximation~\cite{mirsky1960symmetric} for the set of $N$ flattened images with length $m$, $\mathbf{V} \in \mathbb{R}^{N \times m}$ as measured by the Frobenius norm~\cite{eckart1936approximation} $\min \epsilon = \min_{\hat{\mathbf{V}}} \| \mathbf{V} - \hat{\mathbf{V}} \|_F^2 \quad \text{s.t.} \quad \text{rank}(\hat{\mathbf{V}}) \leq n$. Spatial patch-based decompositions do not generally minimize this error~\cite{kersten1987redundancy}.

\ed{Energy compaction is the same statement read from the other side rather than a second property. Because the total energy is fixed, $\|\mathbf{V}\|_F^2 = \|\hat{\mathbf{V}}\|_F^2 + \epsilon$, minimizing the residual is identical to maximizing the retained variance, which the eigenbasis also attains~\cite{hotelling1933analysis} in orthogonal components~\cite{candes2011pca}. Writing the retained power as $\rho = \sum_{i=1}^n \lambda_i$ and the residual as $\epsilon = \sum_{i=n+1}^m \lambda_i$ with eigenvalues $\boldsymbol{\lambda}$, the signal-to-noise ratio of the rank-$n$ estimate~\cite{gonzalez2018digital,jain1989fundamentals}, $\mathrm{SNR} = \rho/\epsilon$, is therefore maximized over all linear approximations of that rank, with $n$ sampled spatial patches~\cite{simoncelli2001natural} attaining neither bound while carrying the parameter burden of spatial attention~\cite{khan2022vitsurvey}. We assume the clinical signal of interest resides on a structured lower dimensional manifold~\cite{tenenbaum2000manifold} and that discriminative information survives in the retained components, which is what makes truncation an efficient tokenization rather than merely a lossy one.}

\subsubsection{Translation invariance}\label{sec:ti}
\ed{Tokens are the magnitudes $s_i = \left| \sum_{j=1}^{m} v_j e^{-\tilde{i} 2\pi \langle \mathbf{k}_i, \mathbf{x}_j \rangle} \right|$ of the complex coefficients, where $\mathbf{k}_i$ is the spatial frequency vector, $\mathbf{x}_j$ the coordinate of the $j^{th}$ pixel and $\tilde{i} = \sqrt{-1}$, ordered by radial frequency $\rho_i = \|\mathbf{k}_i\|$ with $\omega_i = 1/(1+\rho_i)$. Taking the magnitude discards the phase and with it any record of position, so the token sequence is invariant to translation; the cost falls wherever phase carries signal rather than nuisance~(\secname{sec:ti}). The positional embedding $\mathbf{e}_{\mathrm{pos},i}$ then indexes the mode rather than a spatial coordinate, supplying the dimensionality attention requires without reintroducing position.}

\ed{Where the Fourier basis is used, tokens are the magnitudes of the complex coefficients, and a spatial shift of $\mathbf{v}$ multiplies each coefficient by a unit-modulus phase factor. Taking the magnitude discards that factor, so the token sequence is unchanged by translation. The invariance is therefore purchased by discarding phase, and the cost is paid wherever phase carries signal rather than nuisance: a basis should be chosen to match the nuisance structure of the task, which~\secname{sec:oc} and~\secname{sec:ba} bracket from either side.}

\subsubsection{Topology preservation}\label{sec:tp}
\ed{Where the pixel lattice is not the relevant geometry, tokens are projections $s_i = \mathbf{v} \cdot \boldsymbol{\psi}_i$ onto the eigenvectors of the normalized Laplacian $\mathcal{L} = \mathbf{I} - \mathbf{D}^{-1/2}\mathbf{W}\mathbf{D}^{-1/2}$ of a symmetric adjacency $\mathbf{W}$ with degree matrix $\mathbf{D}$, ordered by eigenvalue $0 = \lambda_1 \leq \dots \leq \lambda_n$ with $\omega_i = e^{-\lambda_i \tau}$ for a diffusion scale $\tau$~\cite{belkin2003laplacian}. A low-eigenvalue mode varies slowly across the encoded connectivity, so truncation preserves neighborhood structure rather than smoothness on the grid~(\secname{sec:tp}). On a regular lattice the two coincide and the property is not separable from the preceding one;~\secname{sec:modelnet} separates them on a domain where no cosine basis exists.}

\ed{Where the graph Laplacian is used, the basis vectors are the eigenvectors $\boldsymbol{\psi}_i$ of $\mathcal{L}$ and the ordering is by eigenvalue $0 = \lambda_1 \leq \dots \leq \lambda_n$. The Rayleigh quotient $\lambda_i = \boldsymbol{\psi}_i^{\top}\mathcal{L}\boldsymbol{\psi}_i$ measures how much a mode varies between adjacent vertices, so a low-eigenvalue token is by construction smooth with respect to the connectivity encoded in $\mathbf{W}$ rather than with respect to the pixel lattice. Truncating the sequence therefore discards the modes that oscillate fastest across the graph, which preserves neighborhood structure on domains where the Euclidean grid is not the relevant geometry. On a lattice this property is indistinguishable from~\secname{sec:dctklt}, since the Laplacian eigenvectors of a grid with Neumann boundaries are the cosine basis;~\secname{sec:modelnet} separates them on a domain where no cosine basis exists.}

\ed{A Laplacian basis is fixed only once the graph is: the vertex set, the edge weights $\mathbf{W}$, and which Laplacian is diagonalized. Both the normalized $\mathcal{L} = \mathbf{I} - \mathbf{D}^{-1/2}\mathbf{W}\mathbf{D}^{-1/2}$ and the combinatorial $\mathbf{D} - \mathbf{W}$ give eigenvectors ordered by a Rayleigh quotient that measures variation across edges, so the smoothness reading above holds for either, but they are different bases and the choice must be stated. This paper uses two graphs. In~\secname{sec:modelnet}, the experiment the property claim rests on, $\mathbf{W}$ is the unweighted edge set of one shared icosphere mesh and the combinatorial Laplacian is used; the graph is fixed before any data is seen, identical for every shape, size and fold, so nothing about it is estimated. In Figure~\ref{fig:bases} $\mathbf{W}$ is instead the four-neighbor pixel lattice reweighted by one Voronoi partition drawn from a fixed seed, with the normalized Laplacian; that partition is independent of the phantom and shared by every row, and the fourth row's image is constructed from it, so the panel illustrates the property rather than testing it.}

\ed{Each of the three bases orders tokens by a different notion of what should be kept first: residual variance for PCA, radial frequency for Fourier, and graph smoothness for the Laplacian. Rank is a statistical hierarchy in every case, but the statistic differs, which is why the basis is a modeling choice rather than an implementation detail. Figure~\ref{fig:bases} compares reconstruction under each.}

\subsubsection{Eigenbasis approximation}\label{sec:dctklt}
\ed{The optimality argument above is a property of the eigenbasis, which must be estimated from the training data. The discrete cosine transform supplies most of it without that estimate: for a first-order stationary Markov process the DCT-II asymptotically approaches the Karhunen-Lo\`eve transform as the inter-pixel correlation $\rho \rightarrow 1$~\cite{ahmed1974dct,rao1990dct}. Within the small blocks used here, where correlation between neighboring voxels is high, the fixed basis is therefore close to the optimal one.}

\ed{Three consequences follow, and they are what recommend the block decomposition over a fold-wise PCA in the smallest samples. The basis costs no data, so at $n=25$ nothing is spent estimating it, whereas a PCA basis fit on sixteen training subjects is itself noisy. The basis costs no parameters, as there is no $m \times n$ matrix to store or transmit. And because the transform is fixed rather than refit, out-of-fold scores share a common scale, which a fold-wise PCA does not provide. The DCT is also real-valued, so unlike the magnitude of a Fourier transform it discards no phase, and the cost of that discarding is measured on identical models in~\secname{sec:ba}.}

\subsubsection{Spatial correspondence}\label{sec:corr}
\ed{Each token is a fixed linear functional of the image coordinates: $s_j = \langle \mathbf{e}_j, \mathbf{v} \rangle$ applies the same weights to the same coordinates in every sample. Two consequences bound where any of the bases above can be expected to work. First, $s_j$ is comparable across samples only insofar as coordinate $i$ denotes the same content in each of them, since a single weight $\mathbf{e}_j[i]$ is otherwise applied to unrelated quantities. Second, and more sharply, no informative fixed basis is invariant to translation: if a linear functional satisfies $\langle \mathbf{a}, \mathbf{P}_k \mathbf{v} \rangle = \langle \mathbf{a}, \mathbf{v} \rangle$ for every cyclic shift $\mathbf{P}_k$ and every $\mathbf{v}$, then $\mathbf{P}_{-k} \mathbf{a} = \mathbf{a}$ for all $k$, so $\mathbf{a}$ is constant and the functional is a multiple of the sample mean. The shift-invariant subspace of linear functionals on $\mathbb{R}^m$ is one-dimensional.}

\ed{This is why the invariance of~\secname{sec:ti} is supplied with a magnitude operation rather than a projection: $|\cdot|$ is nonlinear, enabling a translation-invariant token carrying information beyond the mean. It is also why a fixed basis is not a drop-in replacement for a learned tokenization. Patch embedding shares one projection across positions and is therefore approximately translation-equivariant by construction, whereas a fixed basis expresses no such equivariance and instead requires that the data arrive in a common coordinate frame. That frame need not come from registration. Our volumetric cohorts are native-space acquisitions center cropped about the volume midpoint, and the convention of head positioning in the scanner together with a common crop is enough for a given coordinate to denote approximately the same anatomy in every subject; an atlas or a template tightens the correspondence where one is available but is not what establishes it. No comparable convention exists for photographs of objects at arbitrary position, scale and pose, which is the distinction that matters. The stationarity assumed in~\secname{sec:dctklt} is itself a statement about that frame. Correspondence therefore governs how much discriminative signal the leading coefficients carry rather than whether the representation is usable at all. Where it is established the leading components order by content; where it is not they order by position and pose, the discriminative directions move further down the sequence, and a truncation of fixed length retains less of them. The effect appears as a shift in where the crossover falls rather than as an outright failure.}

\subsection{Cost of invariance}\label{sec:magcost}
\ed{Replacing the signed block cosine transform with the magnitude of a block Fourier transform of the same blocks buys invariance to translation~(\secname{sec:ti}) at the price of phase, and on~\secname{sec:ba} that price is measurable. Four model families are run in both bases on identical draws, folds, capacity grids and training budget, so the only difference within a pair of arms is the basis: logistic regression, a multilayer perceptron, and the spectral ViT at one and at two layers. The proposed arm itself is absent from the comparison, because its rank weighting is defined on a frequency-major ordering of the cosine coefficients and has no magnitude counterpart; the two spectral rows are the same encoder without that weighting, which is what makes the contrast attributable to the basis alone. Table~\ref{table:magcost} reports the difference in mean absolute error, per draw, with a paired $t$-test over draws.}

\ed{The magnitude basis costs error in $19$ of the $20$ model-by-size cells and significantly in $11$, by as much as $1.58$ years. The cost is not spread evenly. It falls almost entirely on the two attention encoders, which lose between $0.52$ and $1.58$ years and separate in $7$ of their $10$ cells, while the linear and perceptron heads lose between $0.08$ and $0.88$ years and separate in $4$ of theirs. We read the encoder as the part of the model able to relate signed structure across blocks, and therefore the part with something to lose when phase is discarded. The single cell favoring the magnitude basis is logistic regression at $n=25$, where the ordering reverses in only $7$ of $10$ draws and the paired test is far from significant ($p=0.76$). On the thresholded task of~\secname{sec:hcacls} the magnitude basis is not uniformly worse, logistic regression on it exceeding both the proposed model and its own cosine counterpart at $n=25$; the error metric, which that section gives reasons to treat as primary, is where the bases separate.}

\begin{table*}[tb]
  \caption{\ed{What invariance costs on age estimation. Entries are the Block-$|$FFT$|$ minus Block-DCT difference in mean absolute error in years, averaged over draws, so a positive value means the magnitude basis is worse. Parenthesized values are paired $t$-test $p$-values over draws and bold marks $p<0.05$. Draws number $10$ at $n=25$ and $n=50$ and $5$ above, as in Table~\ref{table:hca-reg}. The four families are those run in both bases; the proposed arm has no magnitude counterpart, for the reason given in the text.}}
  \label{table:magcost}
  \centering
  \begin{small}
  \begin{tabular*}{\textwidth}{@{\extracolsep{\fill}}lccccc@{}}
    \toprule
    Model & $n{=}25$ & $n{=}50$ & $n{=}100$ & $n{=}200$ & $n{=}1000$ \\
    \midrule
    Logistic regression      & $-0.19$ ($0.757$) & $\bf{+0.81}$ ($0.019$) & $\bf{+0.37}$ ($0.042$) & $+0.08$ ($0.603$) & $+0.09$ ($0.312$)  \\
    Multilayer perceptron    & $\bf{+0.88}$ ($0.007$) & $\bf{+0.60}$ ($0.003$) & $+0.33$ ($0.306$) & $+0.19$ ($0.162$) & $+0.29$ ($0.133$)  \\
    Spectral ViT, $1$ layer  & $+1.18$ ($0.072$) & $+1.37$ ($0.052$) & $\bf{+1.58}$ ($<\!0.001$) & $\bf{+0.99}$ ($0.006$) & $\bf{+0.80}$ ($0.002$)  \\
    Spectral ViT, $2$ layers & $+1.41$ ($0.076$) & $\bf{+1.56}$ ($0.031$) & $\bf{+1.16}$ ($<\!0.001$) & $\bf{+1.12}$ ($0.047$) & $\bf{+0.52}$ ($0.001$)  \\
    \bottomrule
  \end{tabular*}
  \end{small}
\end{table*}

\subsection{Age classification}\label{sec:hcacls}
\ed{The same cohort admits a binary label by splitting age at the cohort median ($61$ years, $576$ against $601$), which supplies a ranking metric alongside the error metric of Table~\ref{table:hca-reg}. Results are reported in Table~\ref{table:hca-cls}. We report it separately because the thresholded task saturates: a median split separates the extremes trivially while placing subjects within a few years of $61$ on opposite sides, so a linear ranking of the coefficients suffices and little headroom remains. Logistic regression on the same coefficients is accordingly equal or better from $n=50$ upward, reaching $0.966$ at $n=1{,}000$.}

\ed{We note this is the one task on which a linear head outperforms the proposed model, and that it does so at every size once both bases are counted: on the Fourier magnitude basis logistic regression reaches $0.909$ at $n=25$, above the $0.904$ of the proposed model and the $0.897$ of its own cosine basis counterpart. The task differs from~\secname{sec:sc} in how many coefficients that head receives: the block decomposition supplies a fixed $4{,}096$ to every model here, against at most $128$ components selected by cross-validation on IXI. Reducing that count does not recover the ordering. Repeating the sweep with $512$ and $1{,}728$ coefficients, and with the count chosen by inner cross-validation, leaves the proposed model behind the better linear head at every size but one, by $0.007$ and $0.027$ at $n=50$ and $n=100$ with $512$ coefficients and by $0.029$ and $0.013$ at $n=25$ and $n=50$ when the count is selected rather than fixed; selection on a subsample this small favors whichever setting happens to look best in it. The error metric, which has no such ceiling, is the comparison we treat as primary.}

\begin{table*}[tb]
  \caption{The HCP-Aging cohort with age thresholded at the median. AUC computed on the pooled withheld folds of each subsample, $\pm$ standard deviation over independent subsamples, of which there are $10$ at $n=25$ and $n=50$ and $5$ at the larger sizes. Coefficient-input models are shown on the block cosine basis; the Fourier magnitude variants of the same models are reported in the text. Bold marks the best value in each column and any value a paired test over subsamples does not separate from it.}
  \label{table:hca-cls}
  \centering
  \begin{small}
  \begin{tabular*}{\textwidth}{@{\extracolsep{\fill}}lrccccc@{}}
    \toprule
    Model & Param. & $n{=}25$ & $n{=}50$ & $n{=}100$ & $n{=}200$ & $n{=}1000$ \\
    \midrule
    Logistic regression          & $\bf{4,101}$ & $\bf{0.908 \pm 0.027}$ & $\bf{0.933 \pm 0.028}$ & $\bf{0.935 \pm 0.026}$ & $\bf{0.937 \pm 0.033}$ & $\bf{0.966 \pm 0.003}$  \\
    Multilayer perceptron        & 23,019 & $\bf{0.892 \pm 0.039}$ & $0.913 \pm 0.034$ & $\bf{0.922 \pm 0.019}$ & $0.921 \pm 0.035$ & $0.951 \pm 0.008$  \\
    U-net with attention         & 24,213 & $0.847 \pm 0.077$ & $0.899 \pm 0.032$ & $\bf{0.906 \pm 0.045}$ & $\bf{0.935 \pm 0.013}$ & $0.944 \pm 0.004$  \\
    Swin ViT                     & 23,799 & $0.791 \pm 0.068$ & $0.758 \pm 0.094$ & $0.790 \pm 0.030$ & $0.837 \pm 0.038$ & $0.875 \pm 0.005$  \\
    Spatial ViT                  & 89,492 & $0.757 \pm 0.104$ & $0.751 \pm 0.129$ & $0.766 \pm 0.102$ & $0.849 \pm 0.043$ & $0.912 \pm 0.017$  \\
    \bf{Spectral ViT (proposed)} & 10,715 & $\bf{0.910 \pm 0.036}$ & $\bf{0.922 \pm 0.027}$ & $0.928 \pm 0.028$ & $\bf{0.927 \pm 0.029}$ & $0.947 \pm 0.005$  \\
    \bottomrule
  \end{tabular*}
  \end{small}
\end{table*}

\subsection{Rank weight ablation}\label{sec:rankabl}
\ed{The rank weighting can be omitted or imposed and then learned, and which of these matters depends on how many coefficients it has to act over. Two models share the basis, the tokenization and the encoder, differing only in the rank weights $\boldsymbol{\omega}$. On~\secname{sec:ba}, where the block decomposition supplies $4{,}096$ coefficients, training the weights from their initialization gains between $1.43$ and $2.71$ years of mean absolute error over omitting them, at every size. Both models order identically on~\secname{sec:hcacls}.}

\ed{The exception identifies what is responsible. Repeating the ablation on~\secname{sec:dc}, where the model receives $16$ principal components rather than $4{,}096$ block coefficients, the fixed profile instead helps, by $0.064$, $0.057$, $0.045$ and $0.007$ top-1 across the four sizes ($p \leq 0.032$ throughout), and training the weights from it adds at most a further $0.008$. What differs is the range the profile must decay across: over sixteen coefficients $1/i$ spans a single order of magnitude and reweights mildly, while over four thousand it spans nearly four and suppresses all but the leading few hundred. We therefore read $1/i$ as a reasonable prior whose cost grows with the coefficient count, and the learnable weights as what makes a long coefficient sequence usable rather than as the source of the benefit. Consistently with this, giving each coefficient its own embedding on~\secname{sec:dc} subsumes the explicit hierarchy, an additional $1/i$ pre-scaling slightly degrading it. The property argued in~\secname{sec:lra} is a statement about which coefficients carry variance, not a prescription for how steeply to discount the rest. Initializing at $1/i$ and training is accordingly the configuration we adopt throughout. It is the best of the three at every size on~\secname{sec:ba} and at three of the four on~\secname{sec:dc}, the exception being a $0.002$ difference at the largest digit sample, and it is the only one of the three whose behavior does not depend on how many coefficients the basis supplies: omitting the profile forfeits a real prior, enforcing it is safe only while the sequence is short, and initializing with it recovers the prior without inheriting its sensitivity to length.}

\subsection{Sensitivity to the retained coefficient count}\label{sec:nsens}
\ed{How many coefficients to keep is the one tokenization hyperparameter with no principled setting, so we sweep it on the two cohorts we use for ablation. On~\secname{sec:hcacls} the block decomposition admits $512$, $1{,}728$ or $4{,}096$ coefficients by truncating the per-block cube, and we also report the count chosen by inner cross-validation jointly with the embedding width.}

\ed{Table~\ref{table:nsens} reports the sweep. Performance is not finely tuned to the count. Across an eightfold range the proposed model varies by $0.030$, $0.041$ and $0.041$ pooled AUC at $n=25$, $50$ and $100$, with the full $4{,}096$ we report best or within noise of best at every size. We therefore fix the count and report the sweep.}

\ed{On~\secname{sec:dc} the same sweep over $8$, $16$, $32$ and $64$ principal components, in Table~\ref{table:nsens-mnist}, shows that the best count tracks the sample size: $16$ at $n=50$, $32$ at $n=200$ and $64$ above, with $8$ worst everywhere. The setting we report, $16$, is the best of the four at the smallest sample and the only size at which adding components is harmful, costing $0.025$ top-1 to move to $32$; by $n=5{,}000$ the same move gains $0.009$. This is the expected exchange in miniature, since each additional component must be estimated from the same data, and it is not a capacity effect: under the component-wise embedding, moving from $8$ to $64$ components adds $14{,}336$ weights to a $1.19$ million parameter model, or $1.2\%$. A graph Laplacian model is absent from both sweeps for the reason given in~\secname{sec:modelnet}: on a lattice its eigenvectors are the cosine basis, so it would reproduce the block cosine model rather than provide independent evidence.}

\begin{table}[t]
  \caption{Digit classification on MNIST, top-1 accuracy over $5$ seeds, as the retained principal component count varies. Bold marks the best count at each size; the reported setting is $16$. The spatial baseline, which takes no component count, attains $0.296$, $0.435$, $0.766$ and $0.943$.}
  \label{table:nsens-mnist}
  \centering
  \begin{small}
  \begin{tabular*}{\columnwidth}{@{\extracolsep{\fill}}lcccc@{}}
    \toprule
    Components & $n{=}50$ & $n{=}200$ & $n{=}1{,}000$ & $n{=}5{,}000$ \\
    \midrule
    $8$          & $0.574$      & $0.721$      & $0.798$      & $0.852$ \\
    $\bf{16}$    & $\bf{0.608}$ & $0.773$      & $0.873$      & $0.918$ \\
    $32$         & $0.583$      & $\bf{0.791}$ & $0.884$      & $\bf{0.927}$ \\
    $64$         & $0.584$      & $0.780$      & $\bf{0.885}$ & $\bf{0.927}$ \\
    \bottomrule
  \end{tabular*}
  \end{small}
\end{table}

\begin{table}[t]
  \caption{Age at the median split on HCP-Aging, pooled AUC over independent subsamples, of which there are $10$ at $n=25$ and $n=50$ and $5$ at $n=100$, as the retained coefficient count varies. Standard deviations range from $0.020$ to $0.070$ and are largest at the smallest sample. The final row selects the count inside each outer fold instead of fixing it. Bold marks the setting reported elsewhere.}
  \label{table:nsens}
  \centering
  \begin{small}
  \begin{tabular*}{\columnwidth}{@{\extracolsep{\fill}}lccc@{}}
    \toprule
    Coefficients & $n{=}25$ & $n{=}50$ & $n{=}100$ \\
    \midrule
    $512$             & $0.890$      & $0.890$      & $0.889$ \\
    $1{,}728$         & $0.894$      & $0.915$      & $\bf{0.930}$ \\
    $\bf{4{,}096}$    & $\bf{0.910}$ & $\bf{0.922}$ & $0.928$ \\
    \midrule
    Selected per fold & $0.859$      & $0.893$      & $0.901$ \\
    \bottomrule
  \end{tabular*}
  \end{small}
\end{table}

\subsection{Digit classification}\label{sec:dc}
\ed{We reproduce the simulated crossover on a public multi-class benchmark. MNIST~\cite{lecun1998mnist} digits are subsampled stratified by class at $5$, $20$, $100$ and $500$ images per class, giving $n \in \{50, 200, 1{,}000, 5{,}000\}$, and every model is scored on the full $10{,}000$ image test set. The spectral ViT tokenizes the $16$ leading PCA components, fit on the training subset alone, embedding each coefficient through its own projection weight as in~\secname{sec:ca}; the spatial ViT tokenizes $4\times4$ pixel patches. Both carry the same $128$ dimensional, two layer, four head encoder, which dominates the weight count and balances the two to within $0.4\%$ ($1{,}191{,}450$ against $1{,}195{,}786$ parameters). Neither model's capacity is selected: both run at fixed settings, so the comparison is symmetric, though unlike the volumetric experiments it is not nested. Models were trained for $100$ epochs with a cross-entropy loss using AdamW at learning rate $1\times10^{-3}$ and weight decay $0.01$, and we report the mean over $5$ seeds in Table~\ref{table:mnist}.}

\ed{On~\secname{sec:dc} the crossover reproduces on real images and is located within the grid, as reported in Table~\ref{table:mnist}. The spectral ViT leads by $0.312$ and $0.338$ top-1 accuracy at $n=50$ and $n=200$, in $5$ of $5$ seeds at $p=0.0004$ and $p<0.0001$, and still leads by $0.107$ at $n=1{,}000$ ($p<0.0001$). The spatial ViT overtakes it only at $n=5{,}000$, by $0.025$ top-1 in $5$ of $5$ seeds ($p=0.005$), so the exchange occurs between $1{,}000$ and $5{,}000$ training images. Augmentation is the other standard answer to a small sample, and it acts on the tokenizer rather than beside it. Adding three small affine copies of every training image to both arms lifts the patch tokenizer by $0.069$, $0.200$, $0.157$ and $0.033$ top-1 from $n=50$ to $n=5{,}000$, every size at $p \leq 0.002$, while moving the spectral ViT by at most $0.024$ and the perceptron on the same components by at most $0.009$ in any cell. The gain is specific to a projection that has to be estimated: augmentation supplies the variation a learned tokenizer needs and a fixed one cannot use. The margin accordingly narrows but holds where the claim is made, $0.222$ top-1 and $0.121$ AUC at $n=50$ and $0.111$ and $0.030$ at $n=200$, with the exchange moving to between $n=200$ and $n=1{,}000$. Estimating the basis on the augmented set rather than the clean training split changes no margin by more than $0.009$. A fixed hand-engineered descriptor is the other standard alternative to a learned tokenizer. Fifty-seven first-order, gray-level co-occurrence and Laplacian-of-Gaussian statistics of the same images, the core of a radiomics feature set, reach $0.409$, $0.555$, $0.654$ and $0.697$ top-1 under a linear head and less under a random forest. They lead the patch tokenizer by $0.113$ and $0.120$ at the two smallest sizes and fall behind it from $n=1{,}000$, so a fixed descriptor buys data efficiency of the same kind. The spectral ViT leads them by $0.199$, $0.218$, $0.219$ and $0.221$, a margin flat in the sample where the patch tokenizer's closes: a descriptor fixes the feature set, which cannot grow with the data, where a basis fixes only the projection and admits more coefficients as more become estimable. That the spectral ViT is overtaken once data is plentiful is the expected behavior rather than a failure: $16$ PCA components discard most of the image, which bounds the achievable accuracy, while a patch tokenizer retains everything and needs only enough data to exploit it. The quantity of interest is where the exchange occurs, and on a standard benchmark with a $10{,}000$ image test set it occurs between $1{,}000$ and $5{,}000$ training images.}

\begin{table}[t]
  \caption{Digit classification on MNIST by training set size, over $5$ seeds, evaluated on the full $10{,}000$ image test set. Both models are balanced to within $0.4\%$ of each other in weight count. Bold marks the better value in each pair. The augmented block adds three small affine copies of every training image, $\pm10^{\circ}$, $\pm2$ pixels and $\pm10\%$ scale, to the training split of both arms, leaving the test set untouched; its rows are paired with a re-run of the unaugmented protocol that reproduces the block above to within $0.010$.}
  \label{table:mnist}
  \centering
  \begin{small}
  \begin{tabular*}{\columnwidth}{@{\extracolsep{\fill}}lcccc@{}}
    \toprule
    & \multicolumn{2}{c}{Top-1 accuracy} & \multicolumn{2}{c}{AUC} \\
    \cmidrule(lr){2-3}\cmidrule(lr){4-5}
    $n$ & Spatial & Spectral & Spatial & Spectral \\
    \midrule
    \multicolumn{5}{@{}l}{\emph{No augmentation}} \\
    $50$ & $0.296$ & $\bf{0.608}$ & $0.699$ & $\bf{0.899}$ \\
    $200$ & $0.435$ & $\bf{0.773}$ & $0.810$ & $\bf{0.960}$ \\
    $1{,}000$ & $0.766$ & $\bf{0.873}$ & $0.965$ & $\bf{0.985}$ \\
    $5{,}000$ & $\bf{0.943}$ & $0.918$ & $\bf{0.998}$ & $0.994$ \\
    \addlinespace
    \multicolumn{5}{@{}l}{\emph{Augmented}} \\
    $50$ & $0.362$ & $\bf{0.585}$ & $0.758$ & $\bf{0.879}$ \\
    $200$ & $0.641$ & $\bf{0.751}$ & $0.926$ & $\bf{0.956}$ \\
    $1{,}000$ & $\bf{0.933}$ & $0.866$ & $\bf{0.996}$ & $0.986$ \\
    $5{,}000$ & $\bf{0.976}$ & $0.913$ & $\bf{1.000}$ & $0.994$ \\
    \bottomrule
  \end{tabular*}
  \end{small}
\end{table}

\subsection{Face identification}\label{sec:faces}
\ed{We next test whether the advantage requires medical imaging or only a shared coordinate frame. Labeled Faces in the Wild~\cite{huang2007lfw} supplies natural photographs that are aligned rather than registered, through the unsupervised funneling of~\cite{huang2007funnel}, and aligned faces are the oldest precedent for a global linear basis on photographs. Retaining every identity with at least $20$ images gives $3{,}023$ images over $62$ identities, which we square center-crop and resize to $48^2$ grayscale. Ten images per identity form a fixed, balanced test set of $620$ images, withheld independently of seed, and training draws come from the remainder at $1$, $2$, $3$, $5$ and $10$ images per identity, giving $n \in \{62, 124, 186, 310, 620\}$ for $62$-way identity classification. Tokenization, encoder and training follow the digit experiment exactly; the three trained models balance to within $1.4\%$ ($1{,}198{,}142$ for the spectral ViT against $1{,}214{,}654$ for the spatial ViT and $1{,}197{,}487$ for the perceptron, whose hidden width is set to match). Eigenfaces trains nothing: it is nearest neighbours among the training coefficients, with $k$ selected by inner cross-validation on the training subsample alone. The balanced pool caps the sweep at $10$ images per identity, so this curve brackets the small-sample regime without reaching a crossover.}

\ed{On~\secname{sec:faces} the advantage holds on natural photographs at every size measured, and the four arms separate what the representation contributes from what the encoder does. Against the patch tokenizer, which changes the basis, the spectral ViT leads by $0.047$ top-1 at $n=62$, a single image per identity, rising to $0.146$ at $n=620$. Against the two heads that receive the same $16$ components it leads by less and by a margin that does not grow in the same way: $0.022$ to $0.115$ over eigenfaces, and $0.020$ to $0.032$ over the parameter-matched perceptron. Every one of these fifteen comparisons holds in $5$ of $5$ seeds, at $p \leq 0.045$ and mostly far below (Table~\ref{table:faces}). Most of the gain is therefore the fixed compressive basis, and attention over its coefficients adds a smaller increment that is nonetheless present at every size. That eigenfaces, nearest neighbours in the same subspace, is the weakest of the three coefficient-input arms indicates the components alone do not carry the result. Absolute accuracy is low because the task is $62$-way with as little as one example per class, but the spectral ViT reaches five to fifteen times chance where the spatial ViT reaches two to six. No crossover appears within the grid: the balanced pool allows at most ten images per identity, so the exchange observed on~\secname{sec:dc} lies beyond the reach of this cohort rather than being absent. Faces are aligned but not registered, and the dataset is photographic rather than medical, so the condition that matters is the shared coordinate frame of~\secname{sec:corr} rather than the imaging modality, which~\secname{sec:cifar} confirms from the other direction.}

\begin{table*}[t]
  \caption{Identity classification on Labeled Faces in the Wild by training set size, over $5$ seeds, evaluated on a fixed balanced $620$ image test set of $62$ identities. Eigenfaces, the multilayer perceptron and the spectral ViT all receive the same $16$ principal components and differ only in the head, so the margin over the spatial ViT is the representation and the margin over the other two is the encoder. Eigenfaces is nearest neighbours in that subspace, with $k$ selected by inner cross-validation and resolving to $1$ at every size. Bold marks the best value in each column and any value a paired test over subsamples does not separate from it.}
  \label{table:faces}
  \centering
  \begin{small}
  \begin{tabular*}{\textwidth}{@{\extracolsep{\fill}}lcccccccc@{}}
    \toprule
    & \multicolumn{4}{c}{Top-1 accuracy} & \multicolumn{4}{c}{AUC} \\
    \cmidrule(lr){2-5}\cmidrule(lr){6-9}
    $n$ & Spatial & Eigenfaces & MLP & Spectral & Spatial & Eigenfaces & MLP & Spectral \\
    \midrule
    $62$ & $0.034$ & $0.059$ & $0.061$ & $\bf{0.081}$ & $0.547$ & $0.522$ & $0.612$ & $\bf{0.647}$ \\
    $124$ & $0.044$ & $0.077$ & $0.098$ & $\bf{0.123}$ & $0.568$ & $0.531$ & $0.675$ & $\bf{0.698}$ \\
    $186$ & $0.049$ & $0.081$ & $0.122$ & $\bf{0.150}$ & $0.582$ & $0.533$ & $0.712$ & $\bf{0.739}$ \\
    $310$ & $0.060$ & $0.101$ & $0.159$ & $\bf{0.191}$ & $0.613$ & $0.543$ & $\bf{0.764}$ & $\bf{0.772}$ \\
    $620$ & $0.100$ & $0.131$ & $0.222$ & $\bf{0.246}$ & $0.691$ & $0.559$ & $\bf{0.813}$ & $\bf{0.814}$ \\
    \bottomrule
  \end{tabular*}
  \end{small}
\end{table*}

\subsection{Natural image classification}\label{sec:cifar}
\ed{The benchmarks above are aligned: digits are centered and size-normalized, organ crops are taken about a common anatomy, and our volumetric cohorts, though acquired in native space, are center cropped into approximate anatomical correspondence. CIFAR-10 and CIFAR-100~\cite{krizhevsky2009cifar} are not, so they isolate what~\secname{sec:corr} predicts: objects appear at arbitrary position, scale and pose, and a given pixel index denotes no consistent content across images. We subsample stratified by class to $n \in \{25, 50, 200, 1{,}000\}$ on CIFAR-10 and, because $100$ classes cannot be covered below $n=100$, to $n \in \{100, 200, 300, 500, 2{,}000\}$ on CIFAR-100, scoring on the full $10{,}000$ image test set over $5$ seeds. Protocol, encoder and parameter balance follow~\secname{sec:dc} exactly, with three input channels rather than one.}

\ed{The advantage does not appear at any size, and neither does an outright failure. On CIFAR-10 the two models are statistically indistinguishable at the sizes where the gap elsewhere is largest, $-0.002$ AUC at $n=25$ ($p=0.73$) and $-0.009$ at $n=50$ ($p=0.17$), after which the spatial ViT pulls ahead by $0.026$ at $n=200$ ($p=0.008$) and $0.047$ at $n=1{,}000$ ($p<0.001$). CIFAR-100 behaves the same way, within $0.003$ AUC of the baseline from $n=100$ to $n=300$ and behind by $0.013$ and $0.047$ at $n=500$ and $n=2{,}000$. The perceptron on the same components is ahead of the proposed model on this basis from $n=200$ upward, by $0.030$ and $0.029$ on CIFAR-10 and by $0.015$ to $0.027$ on CIFAR-100, in $0$ of $5$ seeds against it. Attention over these coefficients therefore buys nothing here, which is consistent with the basis itself being the wrong one: where the components are delocalized and correspondence fails, a head that treats them as an unordered vector loses nothing by doing so.}

\ed{The cause is neither the choice of basis nor the truncation. Substituting the basis does not recover the gap: at matched coefficient counts the whole-image and block discrete cosine transforms, the Fourier magnitude, the Fourier coefficient with phase retained, a Haar wavelet, independent components and fixed random projections all trail the PCA basis, which is itself the best of them and only reaches parity. Nor is it the number of coefficients: raising the count from $16$ to $64$ and $128$ changes little, and an model given the complete lossless transform in frequency-ordered tokens of the same width as the baseline's patches does not learn at all, remaining near $0.53$ AUC at every size. Ordering the components by their association with the training labels rather than by variance is no better, costing $0.027$ AUC at $n=50$; the discriminative components are the low variance ones, which few samples estimate least reliably, so the unsupervised ordering is the correct choice under scarcity.}

\ed{What does matter is where the tokens are indexed. A translation permutes the index set of a basis only when its functions are spatially localized, which is what makes a shared projection across tokens meaningful; the principal components of a whole image are maximally delocalized, so no such permutation exists and each coefficient requires its own embedding, making capacity grow with the coefficient count. The patch tokenizer is under no such constraint, which is why it converts additional data into use of its fuller input while a coefficient model is capped. Correspondence enters by determining whether a few coefficients suffice: the first $16$ principal components carry $75.4\%$ of the class-discriminative signal on MNIST against $37.7\%$ on CIFAR-10, where individually discriminative components are found as far down as rank $70$.}

\ed{That diagnosis admits a test, and it passes. A localized block transform taken at a stride smaller than its support is indexed by position, so a shared projection is legitimate. Taken at stride four it is a per-patch invertible map that a learned projection absorbs, and it matches the patch tokenizer accordingly; at stride two it exceeds it by $0.017$, $0.037$ and $0.037$ AUC at $n=25$, $50$ and $200$ ($p=0.045$, $p<0.001$, $p=0.004$). Overcompleteness alone does not account for that, since a patch tokenizer at the same stride is overcomplete in the same way, so we run one: against it the transform leads by $0.029$ AUC at $n=50$ ($p=0.004$) and matches it elsewhere, never falling significantly behind at any size on either dataset. The advantage at $n=25$ is therefore the overlap rather than the basis, and what the fixed transform contributes beyond a learned one of the same shape is confined to the middle of the range. The volumetric experiments already tokenize by block transform at its critically sampled value, so relaxing the stride is the natural next setting of the same tokenizer. The resulting filter bank is fixed and carries no learned parameters, which distinguishes it from a learned convolutional stem~\cite{xiao2021earlyconv}: that line of work reports stability and accuracy gains from stacked strided convolutions at ImageNet scale, where the parameters of the stem are affordable, and does not address the regime of a few tens of examples in which they are not. A fixed transform recovering the same benefit without them is therefore a different claim, and the relevant one here. It also differs from block transform tokenization proposed for inference efficiency~\cite{zhou2026frequencyformer}, which keeps the critically sampled, non-overlapping blocks that we find to be the binding constraint. It is likewise distinct from the line that learns on the coefficients a codec has already produced~\cite{gueguen2018jpeg,ehrlich2019jpegdomain,xu2020freqdomain}. There the block transform is a decoding shortcut: the spatial grid is retained and each frequency becomes a channel of it, so a token remains a position and a coefficient a feature of that position, and the objective is throughput at a scale where some accuracy loss is an acceptable trade. Here the coefficient is itself the token, no grid survives tokenization, and the ordering supplies the embedding hierarchy rather than a bitrate heuristic. The same transform leads at every size on CIFAR-100 as well, by $0.011$ to $0.019$ AUC from $n=100$ to $500$. Its advantage is not uniform in sample size, however. It peaks in the middle of each range and collapses at the largest size of either dataset, to $+0.008$ ($p=0.30$) at $n=1{,}000$ on CIFAR-10 and $+0.004$ ($p=0.22$) at $n=2{,}000$ on CIFAR-100, so the gain is a data efficiency effect of the same kind as the rest of the paper rather than a property of the tokenizer that holds at any scale. Both bases are reported in Table~\ref{table:cifar}. We leave its evaluation on the volumetric cohorts, where the block transform is already in use, to future work. It does not, however, transfer to the aligned benchmarks: on~\secname{sec:dc} the overlapping tokenizer trails the sixteen component basis by $0.312$ top-1 at $n=50$, and on~\secname{sec:faces} by a factor of two at every size, because its several hundred tokens require data that those regimes do not supply. The compact basis is therefore the right tokenizer where correspondence holds and the sample is small, and the overlapping one where correspondence fails, which is the scope the preceding paragraphs set out.}

\begin{table*}[t]
  \caption{Classification on CIFAR-10 and CIFAR-100 by training set size, AUC over $5$ seeds on the full $10{,}000$ image test set. Protocol follows Table~\ref{table:mnist}. Each block is its own paired run, so the spatial column is the patch tokenizer trained alongside that block's spectral model. On the overlapping blocks the patch tokenizer is additionally run at the same stride two, which is the control that separates the fixed transform from the overlap: at stride four the two differ in both, and the critically sampled transform is a per-patch invertible map a learned projection absorbs. The perceptron receives the same coefficients as the spectral ViT and is parameter matched to it. Under the PCA basis these images are not in correspondence, which bounds the advantage as discussed above. Bold marks the best value in each row and any value not significantly different from it; $p$ columns are paired $t$-tests over seeds against each spatial arm.}
  \label{table:cifar}
  \centering
  \begin{small}
  \begin{tabular*}{\textwidth}{@{\extracolsep{\fill}}lcccccc@{}}
    \toprule
    & \multicolumn{2}{c}{Spatial} & & & \multicolumn{2}{c}{$p$ vs spatial} \\
    \cmidrule(lr){2-3}\cmidrule(lr){6-7}
    $n$ & stride 4 & stride 2 & MLP & Spectral & stride 4 & stride 2 \\
    \midrule
    \multicolumn{7}{@{}l}{\emph{CIFAR-10, PCA basis}} \\
    $25$ & $\bf{0.603}$ & -- & $0.586$ & $\bf{0.601}$ & $0.727$ & -- \\
    $50$ & $\bf{0.640}$ & -- & $\bf{0.637}$ & $\bf{0.632}$ & $0.165$ & -- \\
    $200$ & $\bf{0.691}$ & -- & $\bf{0.696}$ & $0.666$ & $0.008$ & -- \\
    $1{,}000$ & $\bf{0.764}$ & -- & $0.747$ & $0.718$ & $<0.001$ & -- \\
    \midrule
    \multicolumn{7}{@{}l}{\emph{CIFAR-10, overlapping block DCT}} \\
    $25$ & $0.602$ & $\bf{0.615}$ & $0.559$ & $\bf{0.618}$ & $0.045$ & $0.522$ \\
    $50$ & $0.640$ & $0.648$ & $0.580$ & $\bf{0.677}$ & $<0.001$ & $0.004$ \\
    $200$ & $0.686$ & $\bf{0.699}$ & $0.630$ & $\bf{0.723}$ & $0.004$ & $0.073$ \\
    $1{,}000$ & $\bf{0.772}$ & $\bf{0.782}$ & $0.679$ & $\bf{0.779}$ & $0.298$ & $0.769$ \\
    \midrule
    \multicolumn{7}{@{}l}{\emph{CIFAR-100, PCA basis}} \\
    $100$ & $\bf{0.588}$ & -- & $\bf{0.591}$ & $\bf{0.588}$ & $0.911$ & -- \\
    $200$ & $0.616$ & -- & $\bf{0.631}$ & $0.616$ & $0.92$ & -- \\
    $300$ & $0.635$ & -- & $\bf{0.649}$ & $0.632$ & $0.435$ & -- \\
    $500$ & $0.664$ & -- & $\bf{0.679}$ & $0.652$ & $0.032$ & -- \\
    $2{,}000$ & $\bf{0.760}$ & -- & $0.734$ & $0.713$ & $<0.001$ & -- \\
    \midrule
    \multicolumn{7}{@{}l}{\emph{CIFAR-100, overlapping block DCT}} \\
    $100$ & $\bf{0.586}$ & $\bf{0.588}$ & $0.542$ & $\bf{0.600}$ & $0.052$ & $0.078$ \\
    $200$ & $\bf{0.621}$ & $\bf{0.623}$ & $0.565$ & $\bf{0.631}$ & $0.053$ & $0.138$ \\
    $300$ & $0.637$ & $0.642$ & $0.582$ & $\bf{0.650}$ & $0.019$ & $0.048$ \\
    $500$ & $0.660$ & $\bf{0.669}$ & $0.601$ & $\bf{0.679}$ & $0.018$ & $0.096$ \\
    $2{,}000$ & $0.761$ & $\bf{0.777}$ & $0.657$ & $\bf{0.765}$ & $0.218$ & $0.092$ \\
    \bottomrule
  \end{tabular*}
  \end{small}
\end{table*}

\subsection{Shape classification on a mesh}\label{sec:modelnet}
\ed{On a regular lattice with Neumann boundaries the graph Laplacian eigenvectors are the discrete cosine basis, so no experiment on images can separate~\secname{sec:tp} from~\secname{sec:dctklt}. A sphere admits no separable cosine basis and is therefore the setting in which the Laplacian is the only natural fixed choice. We reduce each ModelNet40~\cite{wu2015modelnet} shape to its support function $h(\mathbf{d}) = \max_{\mathbf{v}} \langle \mathbf{v}, \mathbf{d} \rangle$ sampled at the $642$ vertices of a thrice subdivided icosahedron, so every object carries a signal on one shared graph and a single Laplacian serves all of them. Per-shape meshes would instead give per-shape eigenvectors with arbitrary sign and unstable ordering, which is the failure of~\secname{sec:corr} in its severest form. Sixteen coefficients are taken from each of five projections, all feeding the same encoder: the Laplacian eigenvectors; a per-fold PCA, which is data dependent and ignorant of the graph; a trainable linear projection estimated end to end, which stands in for a learned tokenizer and differs from the fixed bases only in whether the map is supplied or estimated, where patching the sphere would require an unfolding~\cite{cohen2019icosahedral} and so change the tokenization geometry at the same time; fixed random orthonormal directions, which are data independent and ignorant of the graph; and the signal read at sixteen fixed vertices, which is no projection at all. Shapes are drawn stratified by class to $n \in \{40, \dots, 800\}$ over $40$ classes and scored on the standard $2{,}468$ shape test split over $5$ seeds.}

\ed{Two comparisons separate what the graph contributes from what adaptivity contributes, and the size of each depends on the metric. Against the topology-blind random projection, which differs from the Laplacian in nothing but knowledge of the graph, the Laplacian leads by $0.072$, $0.061$, $0.067$, $0.060$ and $0.026$ top-1 across the five sizes, significant at every one ($p \leq 0.047$); in AUC the same comparison gives $0.023$ falling to $0.004$, so the graph is worth considerably more to the decision than to the ranking. Reading raw vertices is worse still on both. Against PCA the two are indistinguishable at $n=40$, differing by $0.020$ top-1 ($p=0.34$) and $-0.002$ AUC ($p=0.67$) with a per-seed spread several times the difference, after which PCA leads on both metrics from $n=200$. Against the learned projection, which estimates its own tokenizer end to end from the same shapes, the Laplacian leads by $0.071$ and $0.047$ top-1 at $n=40$ and $n=80$ ($p=0.009$ and $p=0.019$), the sizes at which a fixed basis is claimed to pay, and matches it in AUC throughout. The Laplacian is the strongest fixed basis at every size on both metrics. PCA is not ignorant of the graph: the leading sixteen principal components have mean Rayleigh quotient $0.349$ against $0.250$ for the sixteen lowest Laplacian modes and $5.99$ for random directions, and $83\%$ of their energy lies within the span of those sixteen modes, rising to $97\%$ within thirty-two. PCA recovers the topology-adapted subspace from data and then orders within it by variance rather than by eigenvalue. The fixed basis supplies without data what the fitted one spends data to estimate, which is the exchange the sample-size curve measures.}

\ed{Table~\ref{table:modelnet} reports the sweep. Against a learned projection the fixed basis shows the exchange this paper is about, on a domain where no patch tokenizer exists to supply one. The Laplacian leads by $0.071$ and $0.047$ top-1 at $n=40$ and $n=80$ ($p=0.005$ and $p=0.020$), by $0.030$ at $n=200$ ($p=0.063$), and is overtaken between $n=400$ and $n=800$; in AUC the margins are $0.030$ and $0.016$ and the exchange arrives earlier, at $n=200$. At the smallest size the learned projection attains $0.247$ against $0.246$ for a random fixed one: with forty shapes it is not estimated at all, and the $10{,}272$ parameters it carries beyond the fixed bases buy nothing until there is data to fit them.}

\ed{This makes~\secname{sec:tp} and~\secname{sec:dctklt} one principle rather than two properties. A fixed basis that approximates the Karhunen-Lo\`eve transform of the domain costs no data, so it wins while that transform cannot be estimated and loses once it can; the cosine basis does this on a lattice for a first-order Markov process, and the Laplacian does it on a graph for signals smooth over its connectivity. What is domain dependent is the construction of the basis, not the argument for using one.}

\begin{table}[t]
  \caption{Shape classification on ModelNet40 by training set size, top-1 accuracy over $5$ seeds on the standard $2{,}468$ shape test split, with $16$ coefficients and an identical encoder throughout. Chance is $0.025$ over the $40$ classes. The Laplacian and the random projection are both fixed and data independent and differ only in knowledge of the graph; the learned projection estimates its own and carries $10{,}272$ parameters the others do not; the vertex reading is no projection at all. Bold marks the best fixed basis, which is the Laplacian at every size; its advantage over the fitted PCA at $n=40$ is confined to top-1.}
  \label{table:modelnet}
  \centering
  \begin{small}
  \begin{tabular*}{\columnwidth}{@{\extracolsep{\fill}}lccccc@{}}
    \toprule
    $n$ & Laplacian & PCA & Learned & Random & Vertices \\
    \midrule
    \multicolumn{6}{@{}l}{\emph{Top-1 accuracy}} \\
    $40$  & $\bf{0.318}$ & $0.298$ & $0.247$ & $0.246$ & $0.221$ \\
    $80$  & $\bf{0.387}$ & $0.394$ & $0.341$ & $0.327$ & $0.293$ \\
    $200$ & $\bf{0.475}$ & $0.502$ & $0.444$ & $0.408$ & $0.395$ \\
    $400$ & $\bf{0.534}$ & $0.567$ & $0.524$ & $0.474$ & $0.472$ \\
    $800$ & $\bf{0.581}$ & $0.630$ & $0.595$ & $0.555$ & $0.518$ \\
    \midrule
    \multicolumn{6}{@{}l}{\emph{AUC}} \\
    $40$  & $\bf{0.754}$ & $0.756$ & $0.724$ & $0.731$ & $0.720$ \\
    $80$  & $\bf{0.810}$ & $0.820$ & $0.794$ & $0.785$ & $0.778$ \\
    $200$ & $\bf{0.872}$ & $0.892$ & $0.873$ & $0.857$ & $0.855$ \\
    $400$ & $\bf{0.920}$ & $0.931$ & $0.922$ & $0.909$ & $0.907$ \\
    $800$ & $\bf{0.935}$ & $0.950$ & $0.946$ & $0.931$ & $0.927$ \\
    \bottomrule
  \end{tabular*}
  \end{small}
\end{table}

\subsection{Simulated experiments}\label{sec:simdesign}
\subsubsection{Pattern classification}\label{sec:pa}
\begin{figure}[t]
  \centering
  \includegraphics[width=\linewidth]{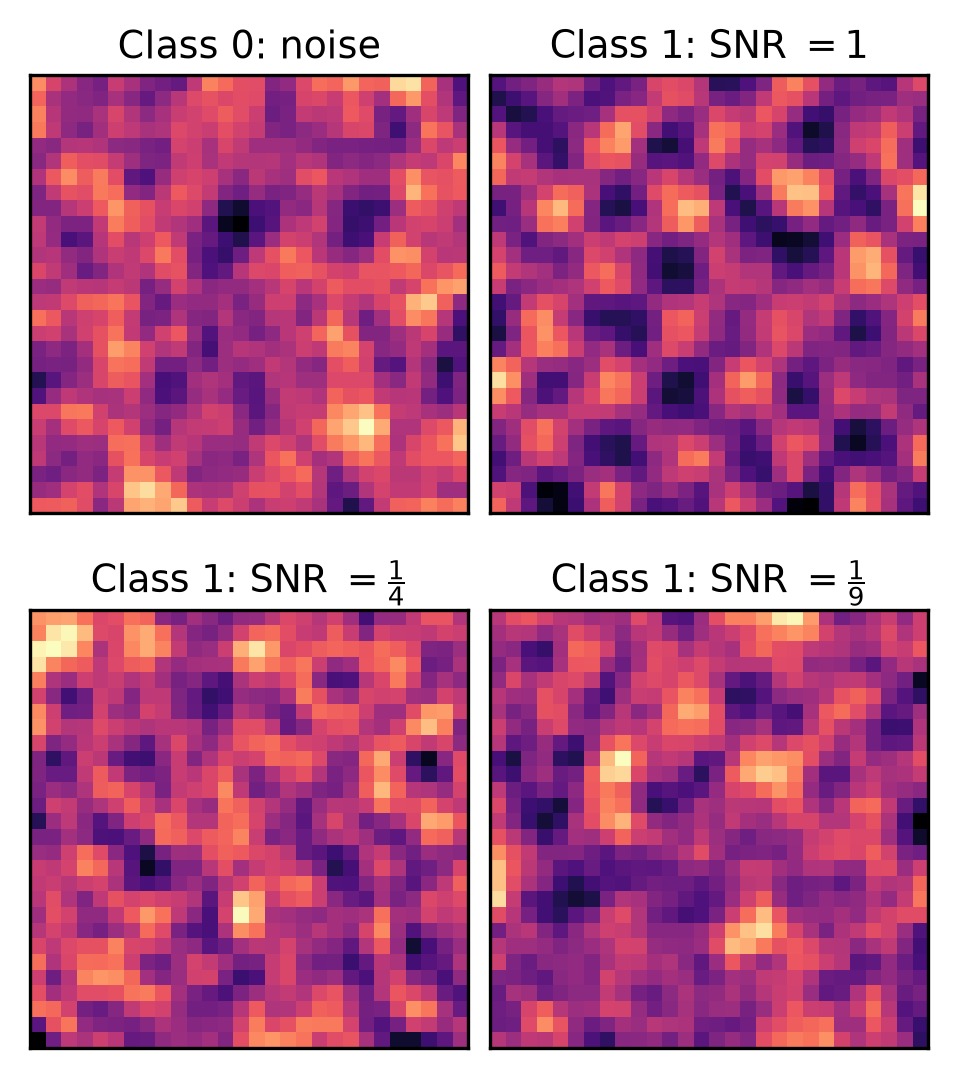}
  \caption{Simulated binary classification between pure noise (class 0) and a checkerboard pattern at decreasing SNR (class 1). The pattern is visible at SNR $=1$ and progressively buried at $\frac{1}{4}$ and $\frac{1}{9}$.}
  \label{fig:sim_viz}
\end{figure}

\ed{We simulate a binary pattern classification task of $N$ images of size $28\times28$ at three signal-to-noise ratios (SNR), $\left[1, \frac{1}{4}, \frac{1}{9}\right]$, in which class 1 contains a checkerboard pattern obscured by Gaussian noise and class 0 contains noise alone, to mimic subtle tissue characteristics accompanying disease progression. Parameter-balanced spectral and spatial ViTs of size $\mathcal{O}(10^3)$ are trained at each $N \in \left[10, 10^2, 10^3, 10^4, 10^5\right]$ over $100$ random seeds. The crossover point $N^*$ is the sample size at which the two achieve comparable performance. The full design is given in~\secname{sec:simdesign}.}
We compare the proposed spectral ViT to a parameter-balanced spatial ViT using a binary pattern classification task by simulating $N$ images of size $28 \times 28$, where $N = \left[ 10, 100, 1000, 10000, 100000 \right]$. To mimic subtle tissue characteristics accompanying disease progression, Class 1 contains a checkerboard pattern obscured by Gaussian noise and Class 0 contains random noise. We select architecture hyperparameters such as the number of components ($16$, spectral ViT, PCA computed on training data only), patch size ($7$, spatial ViT), embedding dimension ($16$), feedforward dimension ($32$), number of heads ($2$), and number of layers ($2$) to balance weight counts between models at size $\mathcal{O}\left(10^3\right)$. Each model is trained on $N$ simulated pairs for $100$ epochs with a binary cross-entropy loss and Adam optimizer with learning rate $1\times 10^{-3}$. Models were tested on $500$ withheld images, and the mean test AUC is reported over $100$ random seeds. We compare model AUCs for datasets simulated with several signal-to-noise ratios (SNR), $\left[1, \frac{1}{4}, \frac{1}{9}\right]$. The crossover point, defined as the sample size $N^*$ where the spatial and spectral ViT achieve comparable performance, is noted. \ed{Example images at each SNR appear in Figure~\ref{fig:sim_viz}.}

\subsubsection{Object classification}\label{sec:oc}
\begin{figure}[t]
  \centering
  \includegraphics[width=\linewidth]{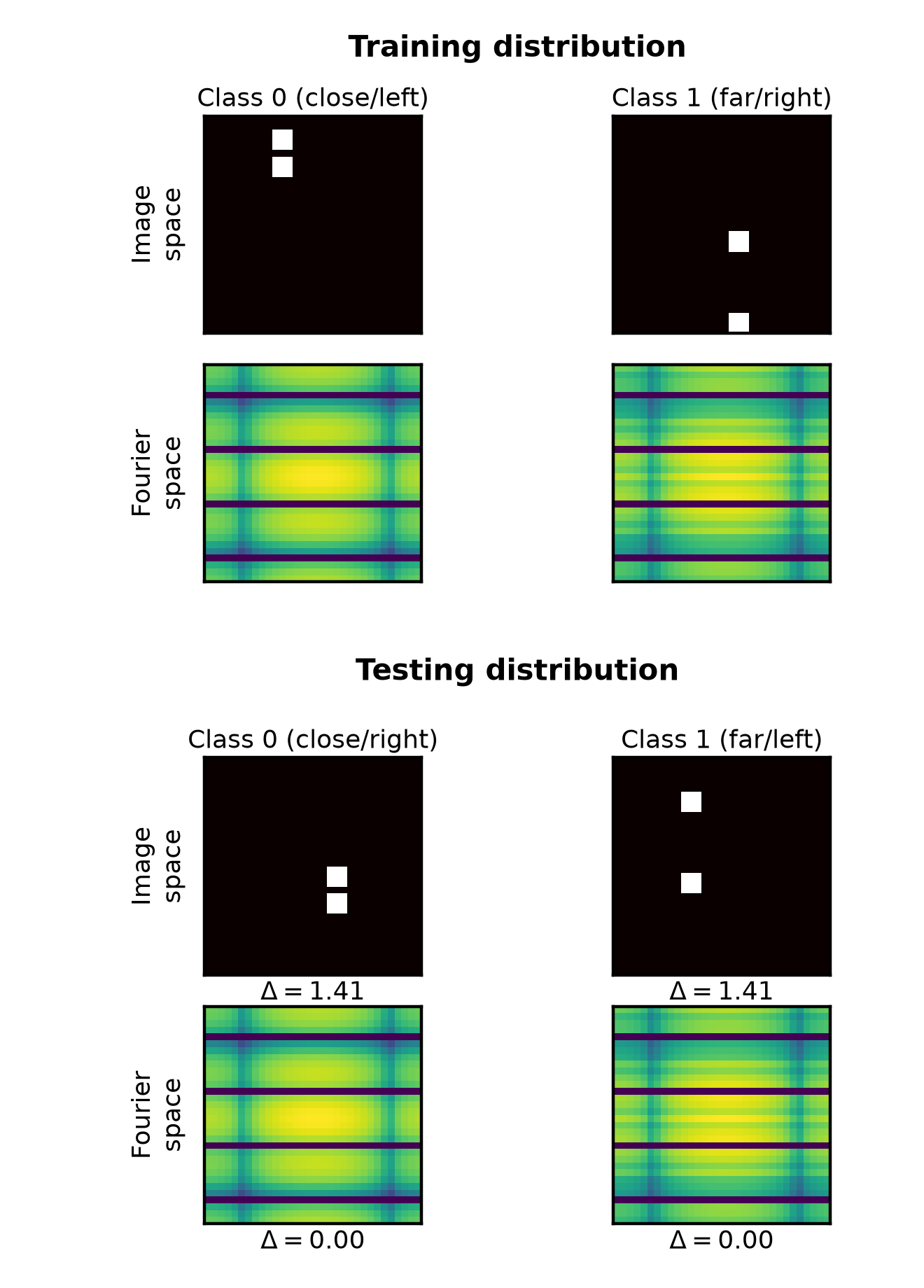}
  \caption{Object classification between close (class 0) and far (class 1) cubes, with position spuriously correlated with the label in training and inverted at test. $\Delta$ gives the normalized difference of each test panel from its training counterpart. In image space the supports are disjoint ($\Delta=\sqrt{2}$, the maximum), while the Fourier magnitude is unchanged to numerical precision ($\Delta=0.00$), since the shift is a pure translation. The two classes remain separated in Fourier space ($\Delta=0.54$ between them), so the basis discards the nuisance while retaining the label.}
  \label{fig:shift}
\end{figure}

\ed{We simulate distribution shift to isolate the translation invariance of the Fourier basis. Two objects are separated by a short distance (class 0) or a long one (class 1), and their position is spuriously correlated with the label during training and inverted at test, so that a model keyed to position scores below chance under the shift rather than merely at it. Networks follow~\secname{sec:pa} and are trained on $10{,}000$ samples over $100$ seeds, scored both on a withheld set from the training distribution and on the shifted set, so that a failure to train is distinguishable from a failure to generalize; the full design is given in~\secname{sec:simdesign}.}
We simulate an example of distribution shift to illustrate the translation invariance of a spectral ViT with Fourier basis. In this example, classes consisting of close objects (class 0) and far objects (class 1) are spuriously correlated in training and testing (shown in Figure~\ref{fig:shift}), with networks similar to those described in \secname{sec:pa}, trained with $10,000$ samples. \ed{Each of $100$ seeds is an independent draw of data and initialization, and every seed is scored twice: on a withheld set from the training distribution and on the shifted set. The first identifies seeds on which a model never learned the task at all, which would otherwise be indistinguishable from invariance in the shifted score.}
We show the spectral ViT outperforms a spatial ViT in limited data regimes for~\secname{sec:pa} in Figure~\ref{fig:sim} of the supplementary material. \ed{At high SNR the spatial ViT achieves comparable performance at $N^*=36$ samples, and at SNR $\frac{1}{4}$ this rises to $N^*=2081$. At SNR $\frac{1}{9}$ no crossover occurs anywhere in the grid: the spectral ViT still leads by $0.053$ AUC at $N=10^5$, on $94$ of $100$ seeds. The advantage conferred by the spectral basis therefore does not simply delay the point at which a spatial model catches up, but survives a four order of magnitude increase in sample size once the pattern is sufficiently buried. Replicating~\secname{sec:oc} over $100$ seeds separates two effects that a single run conflates. The spectral ViT does not always train: it reaches the task in-distribution on $80$ of $100$ seeds, against $100$ of $100$ for the spatial ViT. On the $80$ seeds where both converge, it then classifies the shifted set perfectly on every one ($1.000 \pm 0.000$), while the spatial ViT averages $0.388 \pm 0.166$ and falls below chance on $55$ of them, a systematically inverted rule rather than an uninformative one ($\Delta=0.612$, $80/80$ seeds, $p<10^{-47}$). Invariance under the shift is therefore exact when the model trains, and the residual failure is one of optimization rather than of representation.}

\begin{figure}[t]
  \centering
  \includegraphics[width=\linewidth]{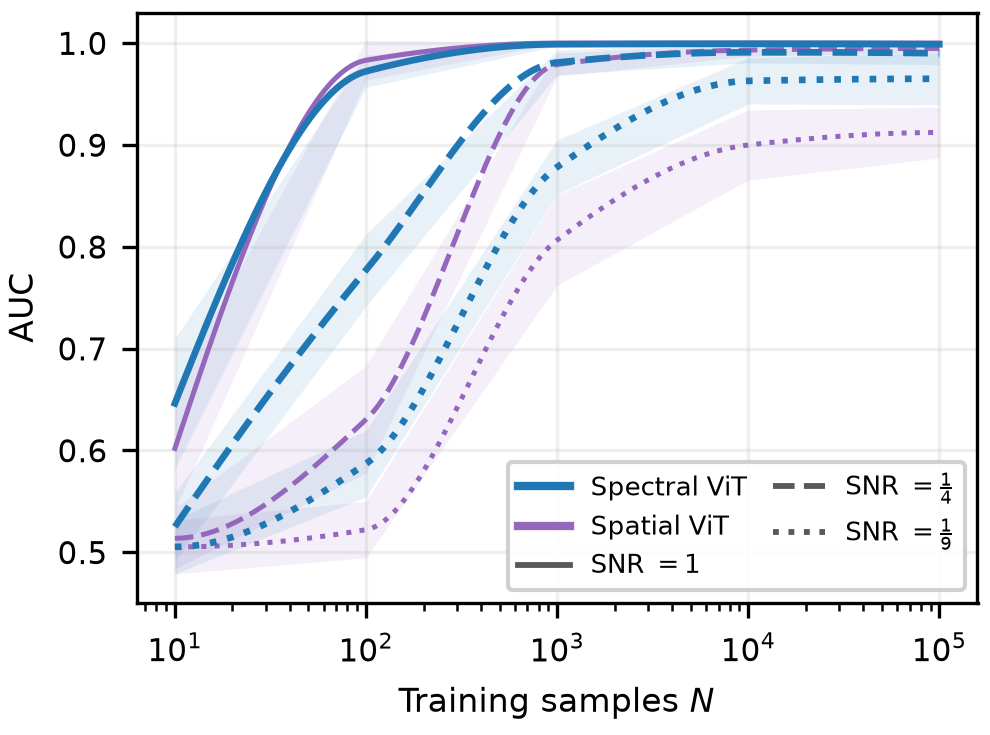}
  \caption{Parameter-balanced spectral and spatial ViT pattern classification performance by sample size $N$ and signal-to-noise ratio (SNR). Color identifies the model and line style the SNR; curves are means over $100$ seeds with $\pm1$ standard deviation bands. The crossover point $N^*$ is the sample size at which the two achieve comparable performance; it moves right as SNR falls, from $N^*=36$ at SNR $1$ to $N^*=2081$ at SNR $\frac{1}{4}$, and at SNR $\frac{1}{9}$ no crossover occurs within the grid.}
  \label{fig:sim}
\end{figure}

\subsection{Paired comparisons on sex classification}\label{sec:ixipaired}
\ed{The draw is the unit of analysis, so each comparison against the proposed model summarizes the independent subsamples by the mean difference in fold AUC, the number of draws in which the proposed model leads, and a paired $t$-test. At $n=25$ it leads every baseline on the mean, by $0.103$ to $0.188$, though in six to nine draws of ten rather than all, and three of the five comparisons separate individually (the spatial ViT at $p=0.013$, logistic regression at $p=0.021$ and the U-net at $p=0.037$); the spread across draws is wide enough at this size that the Swin ViT ($p=0.091$) and the perceptron ($p=0.072$) do not. At $n=50$ it leads all five, by $0.065$ to $0.108$ and in seven to ten draws of ten, and all five separate ($p=0.001$ to $p=0.026$). From $n=100$ the comparison divides. The margin over the two spatial transformers grows and remains significant to the full cohort, reaching $0.111$ and $0.205$ at $n=566$. The coefficient-input models and the U-net instead overtake the proposed model, significantly so at $n=566$, by $0.035$, $0.040$ and $0.053$ AUC ($p\leq0.004$), which is the crossover the main text reports.}

\subsection{Distribution shift}\label{sec:ds}
\ed{We evaluate robustness to distribution shift on the~\secname{sec:sc} task, comparing the spectral ViT to transfer learning from a pretrained model. The simple fully convolutional network (SFCN)~\cite{peng2021sfcn} was pretrained on UK Biobank acquisitions and fine-tuned on IXI, giving it orders of magnitude more data than the spectral ViT sees in total. Both models were trained on uncorrupted images only and saw no noise during training. At test time, Rician noise~\cite{gudbjartsson1995rician}, the appropriate model for magnitude MRI, was added to the withheld fold at signal-to-noise ratios $\left[9, 4, 1, \frac{1}{4}, \frac{1}{9}\right]$, matching those used in~\secname{sec:pa}; the shift is therefore a change in the test distribution rather than a change of site or population. The number of components ($64$) and the fine-tuning schedule were chosen by cross-validation on training data, and a $10\%$ validation split was withheld from each training fold for checkpointing, so the test fold is used only for final scoring. The component count is fixed here rather than selected, and each fold is checkpointed on a withheld validation split, so the uncorrupted AUC of this arm is not comparable with Table~\ref{table:ixi}, where capacity is selected under the one-standard-error rule and training runs to a fixed budget with no checkpointing.}

\begin{figure*}[t]
  \centering
  \includegraphics[width=\linewidth]{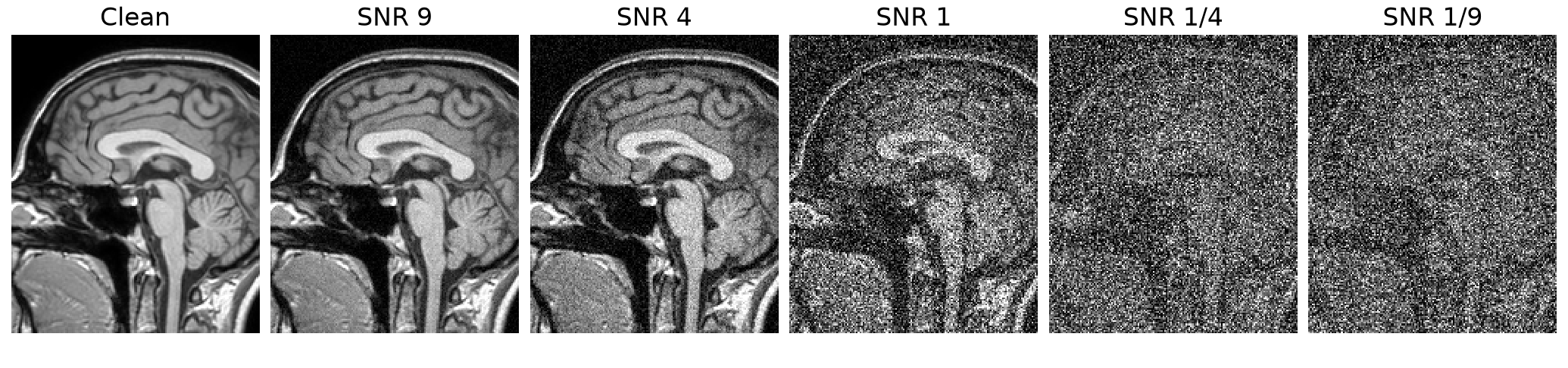}
  \caption{The withheld images at the test-time noise levels of Section~\ref{sec:ds}: one IXI subject under the same Rician noise model at each signal-to-noise ratio, mid-sagittal, on a display window shared across panels. Each corrupted panel is drawn at $75\%$ opacity over the clean slice, so that the anatomy remains discernible as the ratio falls; the models receive the unblended images, in which little of the brain survives below SNR $1$. The rank reversal of Table~\ref{table:shift} falls between SNR $4$ and SNR $1$, and at SNR $\frac{1}{9}$ the spectral ViT still scores $0.747$ AUC against SFCN's $0.454$.}
  \label{fig:snr}
\end{figure*}

\ed{Under distribution shift the two models exchange rank. Transfer learning from UK Biobank leads on uncorrupted images and holds that lead down to SNR $4$, as its pretraining scale would predict; at SNR $1$ and below it collapses to between $0.308$ and $0.479$ AUC, at or below chance, while the spectral ViT maintains AUC above $0.74$ throughout. It is also far more stable once shifted, with a fold-wise standard deviation of $0.007$--$0.054$ against $0.124$--$0.242$. The full sweep is Table~\ref{table:shift}, and Figure~\ref{fig:snr} shows the corrupted inputs. This reproduces the SNR dependence of the crossover characterized in~\secname{sec:pa}, here on real acquisitions and against a pretrained baseline: at SNR $\frac{1}{4}$ and $\frac{1}{9}$ the corrupted volumes retain a voxelwise correlation of only $0.04$ and $0.02$ with the uncorrupted anatomy, so no discriminative structure survives locally and the projection recovers it only by integrating $m$ voxels into each of $n$ coefficients.}

\begin{table*}[t]
  \caption{Robustness to distribution shift in IXI sex classification. Both models are trained on uncorrupted images, with the same Rician noise realization added to the withheld fold of each at test time only. The component count and the fine-tuning learning rate are selected inside each outer fold on its training split alone, so neither model is tuned on a test fold. Mean AUC $\pm$ sample standard deviation over $5$ folds, with a paired $t$-test on the fold-wise AUCs; every row is also consistent in direction across folds, except SNR $4$ where the spectral ViT leads in $1$ of $5$.}
  \label{table:shift}
  \centering
  \begin{small}
  \begin{tabular*}{\textwidth}{@{\extracolsep{\fill}}lllll@{}}
    \toprule
    Test distribution & SFCN (transfer) & Spectral ViT (proposed) & $\Delta$ AUC & $p$ \\
    \midrule
    Uncorrupted        & $\mathbf{0.991 \pm 0.013}$ & $0.931 \pm 0.007$          & $-0.060$ & $0.0016$ \\
    SNR $9$            & $\mathbf{0.991 \pm 0.015}$ & $0.932 \pm 0.009$          & $-0.059$ & $0.0043$ \\
    SNR $4$            & $\mathbf{0.986 \pm 0.021}$ & $0.930 \pm 0.014$          & $-0.056$ & $0.021$  \\
    SNR $1$            & $0.479 \pm 0.242$          & $\mathbf{0.883 \pm 0.008}$ & $+0.404$ & $0.020$  \\
    SNR $\frac{1}{4}$  & $0.308 \pm 0.124$          & $\mathbf{0.795 \pm 0.054}$ & $+0.487$ & $0.0010$ \\
    SNR $\frac{1}{9}$  & $0.454 \pm 0.129$          & $\mathbf{0.747 \pm 0.053}$ & $+0.293$ & $0.015$  \\
    \bottomrule
  \end{tabular*}
  \end{small}
\end{table*}

\subsection{MedMNIST3D classification}\label{sec:medmnist}
\ed{The cohorts of this section are reported separately because the configurations that win on them are not instances of Equation~\ref{eq1}. The selection rule admits a magnitude operation where the spectrum is concentrated or band-limited and a topology operation where the label is binary, and both are nonlinear in the coefficients: a magnitude discards phase, as~\secname{sec:ti} describes, and the topology features are not a projection of $\mathbf{v}$ at all. The fixed, data-independent tokenization and the rank ordering are retained, so the data-efficiency prediction still applies and is what we test here; the linear projection of Equation~\ref{eq1} is not.}

\ed{The volumetric cohorts above are ours, so we repeat the size sweep on a standardized public benchmark where the protocol is fixed by the dataset rather than by us. We use all six cohorts of MedMNIST3D~\cite{yang2023medmnist} at $28^3$: OrganMNIST3D ($971$ training, $610$ test, $11$ organ classes), NoduleMNIST3D ($1{,}158$/$310$), AdrenalMNIST3D ($1{,}188$/$298$), FractureMNIST3D ($1{,}027$/$240$, $3$ classes), VesselMNIST3D ($1{,}335$/$382$) and SynapseMNIST3D ($1{,}230$/$352$), each against a fixed test set rather than a resampled one. We sweep $n \in \{25, 50, 100, 200, 400, 800\}$; the largest rung is $800$ rather than $1{,}000$ because OrganMNIST3D supplies only $971$ training volumes. Classes are unbalanced within cohorts---three organ classes hold roughly $39$ training volumes against $115$ for the rest, and VesselMNIST3D's rarer class $150$---so each size is allocated as evenly as the available counts permit and the remainder spread at random among the classes that still have room.}

\ed{Both models are selected by the same procedure at every sample size and carry the same $128$ dimensional, two layer, four head encoder; the selection rule, the capacity ladder, the draw schedule and the parameter counts are given in~\secname{sec:medmnist-protocol}.}

\ed{The crossover reproduces across all six cohorts and is located within the grid, as reported in Table~\ref{table:medmnist}. Pooled over cohorts the spectral model leads by $0.060$ AUC at $n=25$, significant with the cohort as the unit of analysis ($p=0.036$ over the six), and the margin decays monotonically with the sample---$0.038$, $0.031$ and $0.022$ at $n=50$, $100$ and $200$---before reaching parity at $n=400$ and $n=800$ ($+0.002$, $p=0.79$; $+0.006$, $p=0.67$). The exchange therefore falls between $n=200$ and $n=400$. Counted per cohort the same shape appears: the spectral model wins $6$ of $6$ at $n=25$, $4$ of $6$ at $n=50$ and $n=100$, $3$ of $6$ at $n=200$ and none at $n=400$. This is the data-efficiency claim in its clearest form, and it is a stronger statement than any single sample size would support, since the mechanism---a fixed compressive tokenization buying most where samples are scarcest---predicts the decay rather than merely permitting it.}

\ed{Two cohorts qualify the picture. SynapseMNIST3D reverses: it leads by $0.013$ at $n=25$ and then loses significantly at every larger size, reaching $-0.049$. It is the only cohort that does so, and the pooled margin stays positive through $n=200$ in spite of it. FractureMNIST3D is the one cohort whose advantage does not persist, significant at $n=25$ at $+0.025$ and otherwise within noise of zero.}

\begin{table*}[t]
  \caption{MedMNIST3D by training set size: spectral minus spatial test AUC, positive where the spectral model leads. Each cohort's basis and the baseline's patch size are re-selected at every $n$ by the same inner cross-validation, on the training split alone. $100$ reporting seeds at $n=25$ and $n=50$, $20$ above; paired $t$-test on the per-seed differences. Bold marks a significant win, $\dagger$ a significant loss, both at $p<0.05$. The two models are within $5\%$ smaller to $29\%$ larger of each other in parameter count, varying with the selected patch.}
  \label{table:medmnist}
  \centering
  \begin{small}
  \begin{tabular*}{\textwidth}{@{\extracolsep{\fill}}lcccccc@{}}
    \toprule
    Dataset & $25$ & $50$ & $100$ & $200$ & $400$ & $800$ \\
    \midrule
    Organ & \bf{+.068} & \bf{+.039} & \bf{+.014} & $+.001$ & $-.003^{\dagger}$ & $-.003^{\dagger}$ \\
    Nodule & \bf{+.156} & \bf{+.111} & \bf{+.100} & \bf{+.077} & $+.019$ & $+.015$ \\
    Adrenal & \bf{+.033} & \bf{+.057} & \bf{+.048} & \bf{+.049} & $+.014$ & \bf{+.057} \\
    Fracture & \bf{+.025} & $-.005$ & $+.015$ & $-.001$ & $+.003$ & $+.014$ \\
    Vessel & \bf{+.067} & \bf{+.040} & \bf{+.057} & \bf{+.031} & $+.016$ & $-.010$ \\
    Synapse & \bf{+.013} & $-.016^{\dagger}$ & $-.049^{\dagger}$ & $-.027^{\dagger}$ & $-.036^{\dagger}$ & $-.038^{\dagger}$ \\
    \midrule
    Pooled & \bf{+.060} & \bf{+.038} & \bf{+.031} & \bf{+.022} & $+.002$ & $+.006$ \\
    Cohorts won & $6/6$ & $4/6$ & $4/6$ & $3/6$ & $0/6$ & $1/6$ \\
    \bottomrule
  \end{tabular*}
  \end{small}
\end{table*}

\ed{\paragraph{Occlusion attribution} The figure of the main text is built as follows. With $16$ or $24$ components an input-space gradient for the spectral model occupies a $16$ or $24$ dimensional subspace of $21{,}952$ voxels, so it could only depict the basis; occlusion instead fills a $5^3$ window with the cohort mean, which is exactly zero after the global standardization both models receive, and re-featurizes. Windows are strided by $2$ and each panel is drawn from the resulting $12^3$ grid, the attribution true resolution, interpolated for display. The trained models, the tokenizer and the $n=25$ draw are those of the reported sweep, and each cohort uses the basis and patch its cross-validation selected. The case is chosen by three filters, all symmetric in the two models: both must classify the volume correctly, the weaker model must still respond to occlusion, and the two peak responses must agree in sign. The first two are preconditions for an attribution to mean anything, since a map is uninterpretable where $p(\text{true class})$ is already saturated or already lost; the third keeps the panels from inviting a sign comparison the figure cannot support. Candidates are shortlisted on a coarse pass and resolved on the full one.}

\subsection{MedMNIST3D selection protocol}\label{sec:medmnist-protocol}
\ed{Both models are selected by the same procedure at every sample size, so neither is tuned more than the other. The basis \emph{family} is selected by cross-validation on the training split alone and never consults a test label. Within that family the configuration is re-selected at each $n$ from a fixed ladder of five or six capacities, and the spatial model's patch size is re-selected over $\{4, 7, 14\}$ by the same pooled inner cross-validation. The selector takes the maximum rather than the one-standard-error rule used on the volumetric cohorts. That rule takes the cheapest configuration within one standard error of the best, and at these sample sizes the error is wide enough that most rungs qualify, so it falls through to the smallest: a pull toward parsimony that strengthens as the sample shrinks, which is the wrong direction for a small-sample comparison. The number of selection draws shrinks as $n$ grows so that their product is held near constant, $40$ draws at $n=25$ through $1$ at $n=800$: the selector's total sample budget is fixed, which is the right scaling because inner-fold noise falls as $n$ rises, each of five folds validating on $5$ volumes at $n=25$ and $160$ at $n=800$. Selection draws are disjoint from the draws we report. We report $100$ seeds at $n=25$ and $n=50$ and $20$ above, the reduction following the same argument. We selected the basis and nonlinear operator via cross-validation.}

\ed{Both models carry the same $128$ dimensional, two layer, four head encoder. The spatial model's positional embedding spans one position per patch against the spectral model's one per coefficient, so their parameter counts differ with the patch the cross-validation selects. Across the six cohorts the baseline carries between $5\%$ fewer and $29\%$ more weights than the spectral model ($1{,}238{,}530$ at patch four and $1{,}538{,}690$ at patch fourteen, against $1{,}190{,}402$--$1{,}310{,}347$ for the selected bases), so the comparison is close to matched but is not uniformly conservative: on OrganMNIST3D the baseline is the smaller model.}

\subsection{Float construction}\label{sec:floatdetail}

\paragraph{Construction of the schematic}
The basis vectors $\mathbf{w}_i$ drawn in Figure~\ref{fig:schematic-diagram} are the leading principal components of the $n=1{,}000$ training draw of the digit experiment of~\secname{sec:dc}, each shown at the size of the input it is projected onto, and the input is the first test digit of that experiment. The tinted cells of the figure mark which weights are learned. The spatial ViT learns its linear projection together with one positional embedding per patch, which at patch $4$ is $8{,}448$ weights. The spectral projection is fixed, so the weights the spectral ViT learns on the input side are the rank weights $\omega_i$, initialized at $1/i$, the per-mode projection and the positional embeddings, which is $4{,}112$ weights.

\paragraph{Capacity selection and parameter counts on IXI}
Every model in Table~\ref{table:ixi} has its capacity selected rather than fixed in advance. Inside each outer fold the setting is chosen by $3$-fold cross-validation on that fold's training split alone, under the one-standard-error rule, so that no hyperparameter is informed by the fold it is scored on, and the selection is repeated at every training set size. The selected capacity therefore varies with the sample, and the parameter counts reported in the table are not a single architecture size: each is the mean over the outer folds at the full cohort of $n=566$. The entries themselves are mean fold AUC with the standard deviation taken over the subsample draws, of which there are $10$ at $n=25$ and $n=50$ and $5$ at the larger sizes.

\paragraph{Construction of the basis-family figure}
Figure~\ref{fig:bases} is constructed as follows. The eigenbasis of the empirical column is estimated once from jittered copies of the unshifted phantom and reused throughout, so the lower rows measure how a basis transfers rather than how well it fits. The graphical column diagonalizes the normalized Laplacian of one fixed $60$-cell Voronoi partition, drawn once from a fixed seed and shared by every row, with four-neighbor lattice edges weighted $1$ within a cell and $10^{-4}$ across one; the construction is in~\secname{sec:tp}, and~\secname{sec:modelnet} instead uses the combinatorial Laplacian of an unweighted mesh. Its near-zero cross-region weights make the reconstructions piecewise constant even where the content is not, leaving roughly two of the $128$ retained modes per cell to describe anything inside one, and on an unweighted lattice the construction coincides with the cosine basis, the degeneracy Section~\ref{sec:modelnet} avoids. The bottom row assigns each cell one of six levels so that most cell boundaries are visible rather than merging wherever neighbors share a value. The columns are families rather than operators: the cosine transform, the tokenizer most of the volumetric experiments use, is the same family as the Fourier magnitude with the phase retained, and is not shown separately because two harmonic columns would compare an operation while the others compare families. Scoring the panels by structural similarity rather than PSNR selects the same family in every row.

\paragraph{Cross-site consistent spectral modes}
\ed{The modes shown in Figure~\ref{fig:dgn-diagram} are selected rather than fitted. For each of the $32$ log-magnitude tokens the standardized difference between responders and non-responders was computed separately in each cohort, and a mode retained when both exceeded $0.4$ in the same direction; eight survive, four per nucleus. Each crop is then reconstructed from those modes alone using its own phase, and the class means differenced, so no model enters the computation. The resulting difference map reproduces across sites, at $r=0.67$ and $0.66$ for the higher- and lower-susceptibility nucleus. Because selection uses both cohorts, the cross-site agreement is in part built in, and the figure illustrates where the separable signal lies rather than validating it. Three choices govern how the panels are drawn. Crops are grouped by mean susceptibility rather than hemisphere, as the model's input is. Windows that would leave the volume are centered by edge padding, which affects $10$ of $66$ internal and $4$ of $37$ external cases. The underlay is a cohort average of native-space acquisitions, not an atlas, and the overlay sits at the nucleus centroids measured within it.}

\paragraph{Capacity selection and parameter counts for age estimation.}
The capacity of every model in Table~\ref{table:hca-reg} is selected within each training split rather than fixed across the sweep. Capacity is chosen inside each outer fold, on that fold's training split alone, so the selected setting can differ from fold to fold and from one sample size to the next, and the parameter counts given in the table are means over folds rather than the size of any single fitted model. The number of independent subsamples also differs across the table: $10$ draws at $n=25$ and $n=50$ and $5$ draws at $n=100$, $n=200$ and $n=1{,}000$, so the standard deviations reported in the first two columns and those reported in the remaining three rest on different numbers of draws.

\paragraph{Why occlusion, and the two energy regimes}
\ed{Attribution in Figure~\ref{fig:medmnist-saliency} is by occlusion rather than by gradient because an input-space gradient for the spectral model is confined to the span of its basis and could only depict that basis. The occluded $5^3$ window is filled with the cohort mean and re-featurized through the tokenizer, so the perturbation passes through the basis rather than around it. Across the cohorts the maps separate into two regimes: where the structure is a sparse mask the richest $5^3$ window holds $14$--$20\%$ of the volume energy and both models respond sharply to its removal, whereas on the dense-texture cohorts it holds $0.9$--$2.9\%$ and a global basis is barely perturbed. The maps attribute two different models and none of them adjudicates which is better; the withheld AUC does that. Case selection, the display grid and the per-cohort agreement counts are set out in this section.}

\paragraph{Permutation tests on the external cohort}
The external comparisons of Table~\ref{table:dbs} are tested by permutation, because three external non-responders put any asymptotic test out of range. The spectral ViT's lead on external AUC carries permutation $p=0.043$. Its margin over the clinical covariate models is not resolved at this sample size: the paired permutation test on that difference gives $p=0.24$, and the $95\%$ confidence interval on the AUC difference is $[-0.17,+0.56]$.

\subsection{Training and statistical protocol}\label{sec:protocol}
\ed{The simulated experiments were run on a GeForce RTX 2080 Ti; the volumetric experiments, which carry the capacity selection described with each experiment, were run on a shared institutional cluster across several GPU generations. Throughout we use adaptive moment estimation with parameter-specific learning rates and moment decay rates $\beta_1=0.9$, $\beta_2=0.999$: Adam~\cite{kingma2017adam} for the simulated experiments, and its decoupled weight-decay variant AdamW~\cite{loshchilov2019adamw} with a one-cycle schedule~\cite{smith2019onecycle} for the experiments on IXI and HCP-Aging. These train to a fixed budget rather than selecting a checkpoint, with the exception of~\secname{sec:ds}, where both models early stop on a $10\%$ split withheld from each training fold, following the fine-tuning protocol of the pretrained network being compared against~\cite{peng2021sfcn}; in neither case is a test fold used for anything but final scoring.}

\ed{In~\secname{sec:sc} and~\secname{sec:ba} the budget is $500$ epochs at every sample size, so the number of parameter updates grows with the training split; within a given sample size every model shares the budget. The thresholded variant of the same cohort in~\secname{sec:hcacls} instead holds updates constant at $10^4$ by scaling epochs as $5\times10^4/n$; it is reported separately and no comparison crosses the two. The clinical cohort of~\secname{sec:nscc} is trained for $200$ full-batch steps, a budget chosen to leave predicted probabilities unsaturated, which the quantile-based threshold transfer described there requires.}

\ed{Our choice of statistic follows from how each experiment is partitioned, and in every case the test is paired. Capacity is selected separately inside each cross-validation fold wherever a model is selected at all, so no hyperparameter is informed by a test fold. The benchmark experiments of~\secname{sec:dc},~\secname{sec:cifar} and~\secname{sec:faces} select nothing: every arm runs a fixed configuration and the coefficient-input arms are parameter matched to the spectral ViT, so those comparisons are between representations at equal capacity rather than between selected models. What differs between the remaining experiments is the level at which they replicate. Where a model is refit on repeated subsamples, as in~\secname{sec:sc} and~\secname{sec:ba}, the subsample draw is the unit of analysis: each draw is summarized by its mean across folds and we report a paired $t$-test over the draws together with the number of draws in which the ordering holds. The draws are taken from one cohort and so overlap in subjects, which makes the test anticonservative and the count of draws the more reliable summary. Where a single partition is instead reused across conditions, as in~\secname{sec:ds}, the fold is the unit: each fold carries its own basis and its own selected capacity, which leaves out-of-fold scores on scales that cannot be pooled, so we report a paired $t$-test on the fold-wise AUCs and additionally state the number of folds in which the ordering holds. This test has little power at five folds, and we read a result as supported only when the direction is consistent across them. On the small external cohort of~\secname{sec:nscc}, where three non-responders put any asymptotic test out of range, we use permutation tests. The simulated experiments of~\secname{sec:pa} are summarized by the crossover sample size over $100$ seeds rather than by a test.}

\subsection{Basis families}\label{sec:basisfam}
The framework admits any basis whose coefficients carry an inherent ordering. Here, we outline the \ed{eigenbasis, cosine, Fourier and Laplacian} bases, where each supply a different notion of what a low-rank token should be and inherit specific properties. We demonstrate these in~\secname{sec:basisprops} and illustrate simple reconstructions from tokens in each basis on the Shepp-Logan phantom in Figure~\ref{fig:bases}.

\paragraph{Eigenbasis} Tokens are the coefficients of Equation~\ref{eq1} on a whitened PCA basis, ordered by residual variance. The basis attains the minimum least-squares reconstruction error at any rank and compacts the most signal energy into the fewest components, so truncating the sequence discards the least energy available to discard~(\secname{sec:lra}).

\paragraph{Cosine basis} \ed{Tokens may be the coefficients of a fixed cosine transform, ordered by frequency and requiring no estimate from the training split. For a stationary first-order process the transform approaches the eigenbasis as the correlation approaches one, so it recovers most of the same energy compaction analytically, as shown in~\secname{sec:dctklt}. Applied blockwise it is the tokenizer of~\secname{sec:ba}, and taken at a stride below its support it is localized rather than global, which is the extension proposed in~\secname{sec:corr}.}

\paragraph{Fourier basis} Tokens may also represent the magnitudes of complex Fourier coefficients, ordered by radial frequency. Discarding the phase removes position encoding, so the sequence becomes invariant to translation. The positional embedding then indexes the mode rather than a spatial coordinate. Construction and consequences of this invariance are given in~\secname{sec:ti}.

\paragraph{Laplacian basis} \ed{Where the pixel lattice is not the relevant geometry, tokens may be projections onto the eigenvectors of a graph Laplacian, ordered by eigenvalue. The name denotes a family rather than a single basis: it is determined only once the vertex set and the edge weights $\mathbf{W}$ are given, and $\mathbf{W}$ carries the modeling content, since the eigenvectors are a deterministic function of it. We state $\mathbf{W}$ wherever the basis is used. A low-eigenvalue mode varies slowly across the encoded connectivity, so truncation preserves neighborhood structure rather than smoothness on the grid; the construction is given in~\secname{sec:tp}, and~\secname{sec:modelnet} separates this property from the preceding one on a domain where no cosine basis exists.}

\subsection{Neurostimulation generalization}\label{sec:dbscap}
\ed{On the cohorts of~\secname{sec:nscc} both spatial-domain baselines, the spatial ViT and the U-net with attention, discriminate below chance on internal cross-validation ($0.287$ and $0.409$), so selecting their capacity on that criterion is uninformative: across the grid the spatial ViT's external AUC ranges from $0.794$ to $0.863$ and the U-net's from $0.392$ to $0.686$, while neither reaches a balanced accuracy of $0.53$ at any setting. With three external non-responders the null AUC has a standard deviation of $0.176$, so the U-net's $0.392$ lies $0.6$ standard deviations below chance and is not distinguishable from it. Which of the two capacities the rule selects is itself uninformative, their internal estimates being $0.292$ and $0.239$: the selected one scores $0.392$ externally and the other $0.686$, so the external figure reflects an arbitrary choice between two uninformative options rather than a property of the architecture. This also accounts for the one pairwise comparison that would otherwise appear to resolve. Against each baseline in turn the proposed model's margin is tested by exact permutation on the difference in external AUC, and none reaches significance---$p=0.23$, $0.20$, $0.37$, $0.18$ and $0.46$ against the clinical logistic regression, the clinical transformer, the spectral logistic regression, the spectral perceptron and the spatial ViT---except the U-net at $p=0.046$. That exception is the selected capacity: at the other capacity in the same grid the same comparison gives $p=0.29$, so it turns on a choice the internal criterion could not make.}

\ed{The operating point transferred to the external cohort is a quantile rather than a raw threshold, so the external scores, though never the external labels, enter the decision rule: the threshold applied to one patient is a percentile of all of them. What this can contribute is bounded. It does not supply the specificity, since the quantile is fixed by the internal fit and the models land far apart on it: the spectral ViT reaches $1.000$ at $q=0.47$, calling $17$ of $37$ patients non-responders, where the spatial ViT and the U-net reach the same figure only at $q=0.96$ and $q=0.97$ by calling $35$ of $37$, which is what collapses their balanced accuracy to $0.53$. Nor does it depend on which patients the cohort happens to contain. Refitting the threshold on each of the $37$ leave-one-out subsets moves it between $0.999341$ and $0.999346$, against $0.999343$ on the full cohort; specificity is $1.000$ for every one of the $37$, and balanced accuracy varies between $0.779$ and $0.794$ as a single borderline responder crosses. The AUC in Table~\ref{table:dbs} is threshold-free and does not depend on the transfer at all.}

\ed{The same scarcity quantizes the metrics themselves. With three external non-responders against thirty-four responders the external AUC moves in steps of $1/102$ and specificity in thirds, so a single misranked non-responder shifts AUC by as much as $0.33$ and specificity by one step of three. Events per variable is $0.13$, counting the five internal non-responders against the seven clinical covariates and $32$ spectral tokens, far below conventional guidance for unpenalized models. Every model is therefore regularized, capacity is fixed by internal cross-validation rather than tuned against the external cohort, and a label-permutation test of the full leave-one-out pipeline is uninformative at this granularity.}

\ed{Figure~\ref{fig:dgn-diagram} of the main text asks which spectral coefficients separate the classes consistently at both sites. Eight of the thirty-two modes differ in the same direction in each cohort, four per nucleus, and reconstructing each crop from those modes alone gives a responder minus non-responder difference that reproduces across sites ($r=0.67$ and $0.66$ for the higher- and lower-susceptibility nucleus). We report this in place of a saliency map because no model enters the computation, so the agreement is a property of the cohorts rather than of a fitted attribution. The modes are selected for cross-site consistency, so the figure illustrates where the separable signal lies rather than validating it, and with five internal and three external non-responders we do not advance any single mode as a biomarker. The retained set excludes the constant mode of either nucleus, so what separates the classes is the arrangement of susceptibility within the nuclei rather than their mean, consistent with accounts implicating tissue organization alongside iron content in Parkinson's severity~\cite{roberts2025cp}. Within that set the effect does not track spatial scale: two modes at the same radial frequency carry opposite signs ($+0.85$ and $-0.82$), and the discriminating structure is instead directional.}

\ed{As reported in Table~\ref{table:dbs}, relative to the clinical model, the spectral ViT rectifies $11$ of the $37$ external patients ($1$ non-responder and $10$ responders) while misclassifying $4$ that the clinical model placed correctly, raising the number correctly classified from $16$ to $23$.}